%% file: main.tex
\documentclass[11pt]{article}

\usepackage[preprint]{acl}

\usepackage{times}
\usepackage{latexsym}
\usepackage[T1]{fontenc}
\usepackage[utf8]{inputenc}
\usepackage{microtype}
\usepackage{inconsolata}
\usepackage{url}
\usepackage{booktabs}
\usepackage{amsmath}
\usepackage{amsfonts}
\usepackage{amssymb}
\usepackage{nicefrac}
\usepackage{xcolor}
\usepackage{colortbl}
\usepackage{pifont}
\usepackage{enumitem}
\usepackage{graphicx}
\usepackage{multirow}
\usepackage{fvextra}
\usepackage[most]{tcolorbox}
\usepackage{cuted}

\definecolor{promptblue}{RGB}{7, 10, 170}
\definecolor{promptteal}{RGB}{0, 104, 115}
\definecolor{promptviolet}{RGB}{92, 45, 145}
\definecolor{promptorange}{RGB}{168, 86, 0}
\definecolor{promptmagenta}{RGB}{163, 20, 96}
\definecolor{promptgray}{RGB}{70, 70, 70}

\newcommand\blfootnote[1]{%
  \begingroup
  \renewcommand\thefootnote{}\footnote{#1}%
  \addtocounter{footnote}{-1}%
  \endgroup
}

\newcommand{\gain}[1]{{\tiny\textcolor{green!45!black}{~(+#1)}}}

\newcommand{\abldrop}[1]{{\tiny\textcolor{green!45!black}{~(-#1)}}}

\newtcolorbox{promptbox}[2]{
  enhanced,
  colback=#1!5!white,
  colframe=#1,
  colbacktitle=#1,
  boxrule=1pt,
  arc=5pt,
  outer arc=5pt,
  left=7pt,
  right=7pt,
  top=7pt,
  bottom=7pt,
  title={#2},
  fonttitle=\bfseries\small,
  coltitle=white,
  titlerule=0pt,
  toptitle=4pt,
  bottomtitle=4pt,
  lefttitle=7pt,
  righttitle=7pt,
  before skip=10pt,
  after skip=10pt
}

\title{OmniReasoner: Thinking with Long Audio-Video via Native Tool Use}

\author{
  \textbf{Yu Chen\textsuperscript{1,2,*}},
  \textbf{Caorui Li\textsuperscript{3,*}},
  \textbf{Ziyu Xiong\textsuperscript{3}},
  \textbf{Yidong Wang\textsuperscript{4}},
  \textbf{Mingqi Gao\textsuperscript{5}},
\\
  \textbf{Shuman Liu\textsuperscript{4}},
  \textbf{Biao Liu\textsuperscript{3}},
  \textbf{Chunfeng Yang\textsuperscript{3}},
  \textbf{Anxiang Zeng\textsuperscript{4}},
  \textbf{Haibo Zhang\textsuperscript{4,\dag}},
  \textbf{Chaofan Chen\textsuperscript{6,\dag}}
\\
\\
  \textsuperscript{1}University of Chinese Academy of Sciences,
  \textsuperscript{2}Institute of Automation, CAS,
  \textsuperscript{3}Southeast University,
\\
  \textsuperscript{4}Shopee,
  \textsuperscript{5}Tsinghua University,
  \textsuperscript{6}Beijing University of Technology
\\
  \small{
    \href{mailto:yu.chen.8525@gmail.com}{yu.chen.8525@gmail.com}
  }
}

\begin{document}

\maketitle

\blfootnote{\textsuperscript{*}Equal contribution. Yu Chen's work was done during an internship at Shopee.}
\blfootnote{\textsuperscript{\dag}Corresponding authors.}

\begin{abstract}
Long audio-video reasoning is difficult for omnimodal LLMs because the decisive evidence is often sparse, cross-modal, and too expensive to preserve with uniformly high-fidelity inputs. We introduce \textbf{OmniReasoner}, a tool-use post-training framework for \emph{Thinking with Long Audio-Video}: omni-modal LLMs learn, via supervised fine-tuning and reinforcement learning, to decide whether and where to call a zoom-in tool before answering. OmniReasoner first builds a low-cost global preview of the full stream and then, when needed, calls the zoom-in tool with a requested temporal interval for higher-fidelity visual and audio inspection before answering. Because the model observes different sampling granularities before and after this call---a sparse global preview and a denser local clip---we introduce \emph{TimeAnchor}, which keeps the tool's temporal argument valid and round-trip-consistent across these granularities, rather than tied to frame indices from a particular sampling rate. To make this tool-use behavior trainable without expensive manual interval annotation, we build a \emph{Temporal Augmented Data Engine} that synthesizes tool-use post-training trajectories by video editing and composition. Experiments across omnimodal and video benchmarks show that OmniReasoner improves both answer accuracy and temporal grounding while concentrating high-fidelity computation on informative regions. Code is available at \url{https://github.com/RockyChen0205/OmniReasoner}.
\end{abstract}

\section{Introduction}
\label{sec:intro}

Long audio-video reasoning requires omni-modal Large Language Models (LLMs)~\citep{xu2025qwen3,ye2025omnivinci,fu2025vita} to connect evidence that is sparse in time and distributed across modalities. In an hour-long recording, the decisive clue may be a brief visual action, a short spoken phrase, a background sound, or the temporal coincidence between what is seen and what is heard. Preserving every such scattered clue at high fidelity is prohibitively expensive for omni-modal LLMs: dense video frames already consume substantial context, while audio introduces a continuous token stream on top of the visual sequence. Uniform downsampling keeps broad coverage but often removes the fine-grained evidence that determines the answer.

Recent long-video methods address this uniformity dilemma by moving beyond one-shot uniform perception. Agentic systems search for relevant frames or segments \citep{wang2024videoagentlongformvideounderstanding}, native tool-calling methods crop and re-inspect promising clips \citep{yang2025longvtincentivizingthinkinglong}, adaptive zoom-in models trade global coverage for local resolution \citep{fu2025lover1advancinglongvideo}, and reinforcement-learning approaches optimize temporal search strategies \citep{pan2025timesearchradaptivetemporal}. These works show that long-video reasoning benefits from selective evidence acquisition. However, they are still largely designed around visual evidence: \textit{the model learns where to look, but how should it jointly decide what to look at, what to listen to, and when the interaction between the two modalities is decisive?}
% However, they are still largely designed around visual evidence: the model decides where to look, but not how to jointly decide what to look at, what to listen to, and when the interaction between the two modalities is decisive.

% Prior omni-modal and audio-video reasoning works attack the complementary problem of cross-modal understanding.

\begin{figure*}[t]
    \centering
    \includegraphics[width=\textwidth]{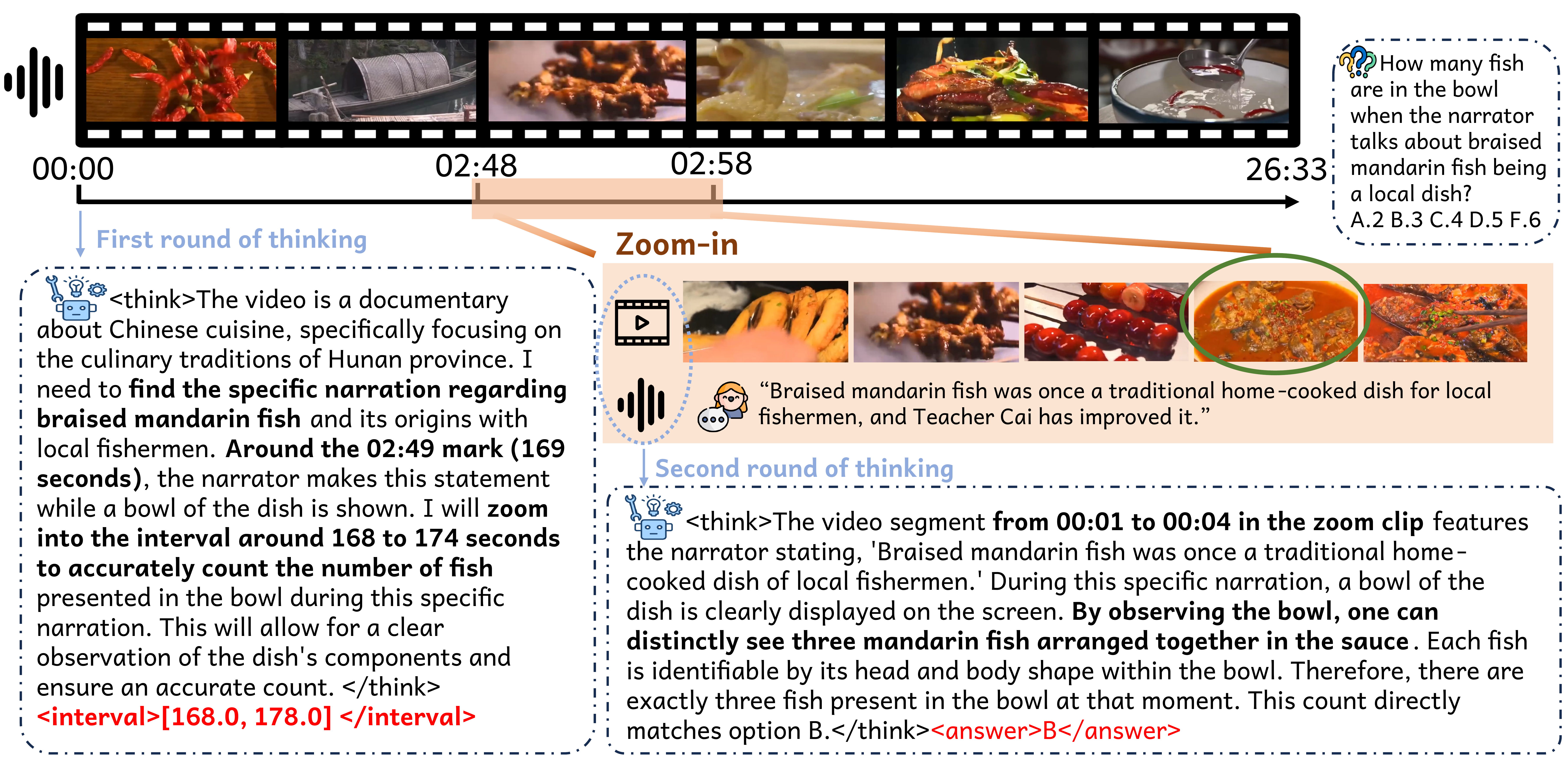}
    \caption{\textbf{An OmniReasoner inference example.} The question asks how many fish are in the bowl when the narrator talks about braised mandarin fish as a local dish. OmniReasoner first identifies the relevant narration around the 02:49 mark and requests a zoom-in interval around that moment. In the returned clip, the narration and the bowl are observed together, allowing the model to count three mandarin fish. The final answer is therefore option B.}
    \label{fig:intro_omnireasoner_example}
\end{figure*}

Recent omni-modal and audio-video reasoning works have begun to explore joint perception across sight and sound, yet they frame the problem from a different angle. Omni-R1 decomposes global video-audio reasoning and fine-grained grounding with a two-system RL framework \citep{zhong2025omnir1reinforcementlearningomnimodal}; EchoInk-R1 and OmniVideo-R1 use reinforcement learning to improve audio-visual reasoning behavior \citep{xing2025echoinkr1exploringaudiovisualreasoning,chen2026omnivideor1reinforcingaudiovisual}; LatentOmni explores unified audio-visual latent reasoning for omni-modal understanding \citep{dai2026latentomni}; and OmniGAIA studies native omni-modal agents with external tools and active perception \citep{li2026omnigaiatowardsnativeomnimodal}. These advances demonstrate that omni models need more than stronger perception. Yet the central training target is usually mixed-modality reasoning, temporal localization, or open-world tool use, rather than the media-internal question of how a model should spend limited high-fidelity computation inside a long audio-video stream.

% Humans rarely solve this problem by inspecting every frame and every sound at full fidelity. We skim the video and audio to form a rough timeline, then drag the progress bar back to suspicious moments and look or listen more carefully. This is not merely a human analogy; it is a computational strategy for allocating expensive perception only where evidence is likely to be useful. It suggests that long audio-video reasoning should be formulated as \emph{adaptive evidence acquisition}: the model should decide whether more evidence is needed, where to retrieve it, and how to use it for final reasoning. In short, OmniReasoner teaches omni models to decide when and where to look and listen again in long audio-video.

% \begin{figure*}[t]
%     \centering
%     \includegraphics[width=\textwidth]{figures/intruduction.pdf}
%     \caption{\textbf{An OmniReasoner inference example.} The question asks how many fish are in the bowl when the narrator talks about braised mandarin fish as a local dish. OmniReasoner first identifies the relevant narration around the 02:49 mark and requests a zoom-in interval around that moment. In the returned clip, the narration and the bowl are observed together, allowing the model to count three mandarin fish. The final answer is therefore option B.}
%     \label{fig:intro_omnireasoner_example}
% \end{figure*}

Humans rarely solve this problem by inspecting every frame and every sound at full fidelity. Instead, we skim the video and audio to form a rough timeline, then drag the progress bar back to suspicious moments and look or listen more carefully. This is not merely a human analogy, but a computational strategy that suggests post-training omni models with an explicit zoom-in tool: the model should decide whether more evidence is needed, where to point the tool, and how to use the returned evidence for final reasoning. In short, we seek to teach omni models, through post-training, to decide when and where to look and listen again in long audio-video.

In this paper, we introduce \textbf{OmniReasoner}, a tool-use post-training framework for \emph{Thinking with Long Audio-Video}. OmniReasoner first consumes a low-cost global preview of the full stream and either answers directly or calls a zoom-in tool with a requested temporal interval for closer inspection. The system then retrieves a higher-fidelity audio-video clip for the requested interval, and the model answers conditioned on both the global context and the zoomed evidence. Temporal zoom-in is therefore not a fixed preprocessing rule but a learned tool-use behavior: the post-training objective is to teach the model not only how to answer, but also when to call the tool and where it should be pointed. Figure~\ref{fig:intro_omnireasoner_example} shows a concrete inference example.

This formulation raises a tool-argument grounding issue that is specific to long audio-video tool use. Prior work has shown that explicit temporal markers, frame numbers, and time-aware representations can improve video temporal localization \citep{chen2024timemarkerversatilevideollmlong,Wu2024,zhang2026timelensrethinkingvideotemporal,chen2025chronusomniimprovingtimeawareness}, but that work operates within a single, fixed-fidelity video. Our setting is different: the model moves between different sampling granularities across one tool call---a sparse global preview before calling the zoom-in tool, and a dense local clip after. A frame-index argument computed under one sampling grid does not resolve to the same moment under the other, and audio does not naturally follow visual frame numbering either. Inspired by how media systems such as \texttt{ffmpeg} align audio and video using absolute time, we introduce \emph{TimeAnchor} so that the zoom-in tool's temporal argument stays valid across this boundary: a call such as ``inspect 32--38 seconds'' resolves to the same moment whether the model is viewing the sparse preview, the dense zoomed clip, or the corresponding audio segment. To make this tool-use behavior trainable at scale, we further build a \textbf{Temporal Augmented Data Engine} that generates tool-use post-training trajectories through editing and composition, providing supervision for when to call the tool, where to point it, and how to answer from retrieved evidence.

Our main contributions are:

% \begin{itemize}
\begin{itemize}[leftmargin=*, label=]
\item \ding{182} We propose \textbf{OmniReasoner}, a tool-use post-training framework for long audio-video reasoning, where omni models learn, via SFT and RL, to decide \emph{whether} and \emph{where} to call a zoom-in tool before answering.
\item \ding{183} We introduce \textbf{TimeAnchor}, which keeps the zoom-in tool's temporal argument valid and round-trip-consistent across the different sampling granularities the model observes before and after the tool call, rather than tied to a particular frame grid.
\item \ding{184} We build a \textbf{Temporal Augmented Data Engine} and a practical post-training recipe that synthesize tool-use trajectories at scale, providing supervision for when to call the zoom-in tool, where to point it, and how to answer from retrieved evidence.
\end{itemize}

\section{Related Work}
\label{sec:related}

\input{related_work.tex}

\section{Method}
\label{sec:method}

\subsection{OmniReasoner Overview}
\label{sec:method_overview}

\begin{figure*}[t]
    \centering
    \includegraphics[width=\textwidth]{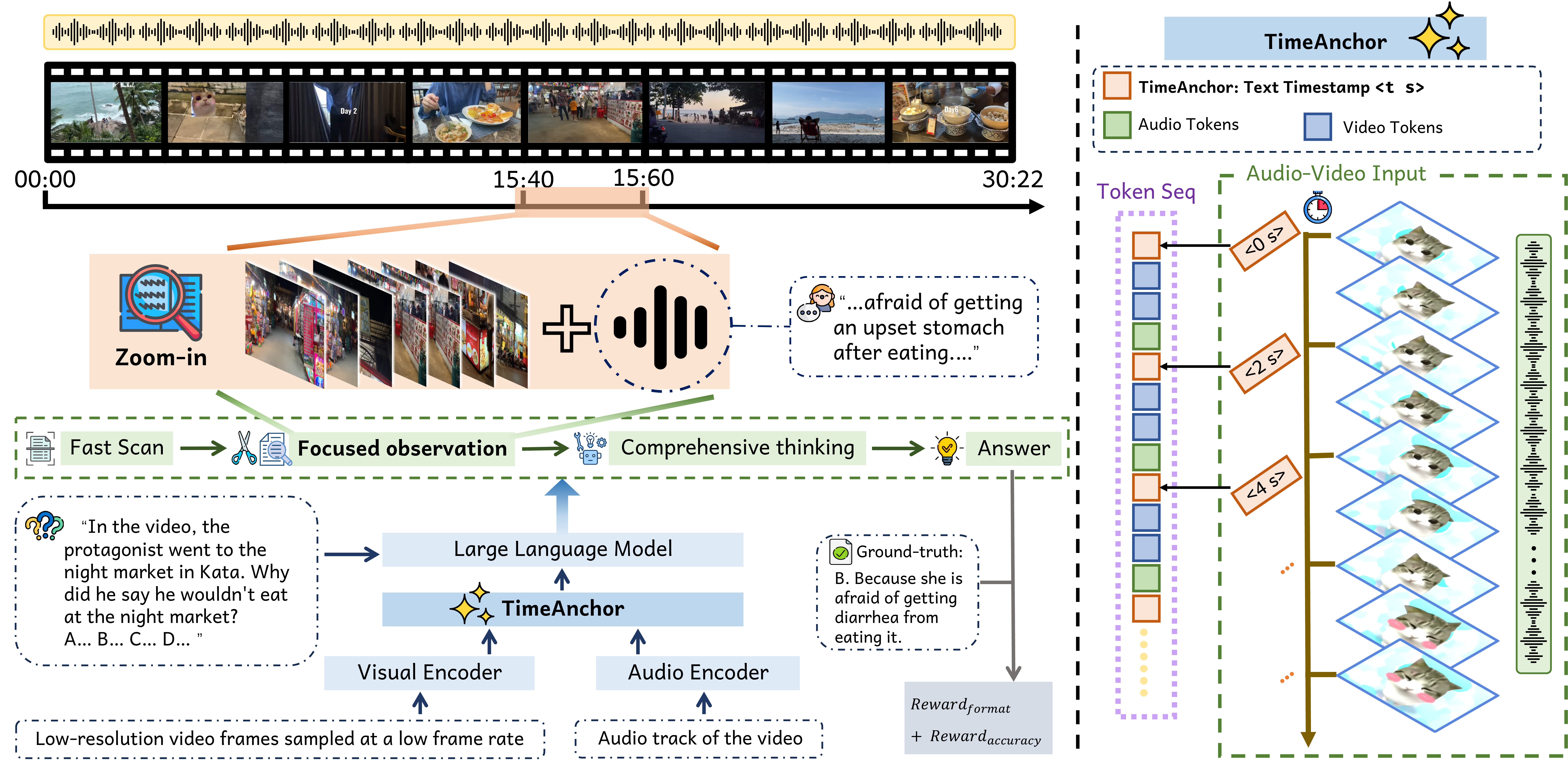}
    \caption{\textbf{Overview of OmniReasoner.} \emph{Left:} the inference workflow. A visual encoder consumes low-frame-rate frames and an audio encoder consumes the full audio track; from this low-cost global scan and the question, OmniReasoner either answers directly or requests a TimeAnchor-grounded interval for focused observation, and the returned higher-fidelity local audio-video clip drives comprehensive thinking and the final answer, with reinforcement learning optimizing the policy under format and accuracy rewards. \emph{Right:} how TimeAnchor forms the audio-video token sequence. Following Qwen-Omni's time-interleaving, the stream is split into two-second chunks by absolute time, and within each chunk the video tokens precede the audio tokens; TimeAnchor simply prepends a plain-text absolute-time marker \texttt{<t seconds>} to every chunk, so a requested interval \texttt{[s,e]} resolves to the same moment across the sparse global preview and the dense local clip.}
    \label{fig:method_omnireasoner}
\end{figure*}

OmniReasoner takes a long audio-video input $x$ of duration $T$ and a question $q$ as input. Long audio-video reasoning involves a tension between temporal coverage and perceptual fidelity. The answer may depend on a short visual action, a brief speech segment, a sound event, or an audio-visual coincidence, while processing the full stream at high fidelity wastes context on irrelevant moments. OmniReasoner formulates this as a tool-use decision problem over the original timeline, as illustrated in Figure~\ref{fig:method_omnireasoner}.

% \begin{figure*}[t]
%     \centering
%     \includegraphics[width=\textwidth]{figures/method-omnireasoner.pdf}
%     \caption{\textbf{Overview of OmniReasoner.} Given a long audio-video input and a question, OmniReasoner first performs a low-cost global scan, then either answers directly or requests a TimeAnchor-grounded interval for focused observation. The retrieved local audio-video evidence supports the final answer, while TimeAnchor keeps zoom-in actions aligned with the original timeline across sampling granularities.}
%     \label{fig:method_omnireasoner}
% \end{figure*}

The model first constructs a global observation
\begin{equation}
    g = \Phi_{\mathrm{global}}(x),
\end{equation}
which preserves the full temporal structure of the input at a low perceptual cost. Conditioned on $g$ and the question $q$, the policy makes a tool-use decision:
\begin{equation}
\begin{aligned}
    a_1 &\sim \pi_\theta(\cdot \mid g, q), \\
    a_1 &\in \{\operatorname{answer}(y), \operatorname{zoom}([s,e])\}.
\end{aligned}
\end{equation}
where $0 \leq s < e \leq T$. The answer action is used when the global observation provides sufficient evidence. The zoom action selects a candidate evidence interval on the original timeline.

For a selected interval $[s,e]$, the media environment returns a local evidence observation
\begin{equation}
    z_{s:e} = \Phi_{\mathrm{local}}(x,[s,e]),
\end{equation}
which exposes the corresponding audio-video content at higher fidelity. The final answer is generated from the question, the global observation, and the retrieved local evidence:
\begin{equation}
    y \sim \pi_\theta(\cdot \mid g, q, z_{s:e}).
\end{equation}
The global observation supports temporal reasoning over the complete input, and the local observation provides fine-grained audio-visual evidence around a selected moment.

The zoom-in tool's requested interval must denote the same moment when the model moves from the sparse global observation to the dense local clip. TimeAnchor keeps this temporal argument consistent across the tool call.

\subsection{TimeAnchor for Audio-Video Grounding}
\label{sec:timeanchor}

Qwen-Omni encodes an audio-video stream with a time-interleaving scheme: it splits the stream into two-second chunks by actual time and, within each chunk, places the video tokens before the audio tokens. Time is thus carried only implicitly, by a chunk index defined relative to the frames that are actually sampled. This is exactly what breaks across a zoom call: when the model moves from the sparse global observation to a densely sampled local clip, the same physical moment falls under a different sampling grid, so a raw chunk or frame index no longer points to the same place. Yet the zoom-in tool takes a temporal interval as its argument, and that argument must survive this change of granularity.

\begin{figure*}[t]
    \centering
    \includegraphics[width=\textwidth]{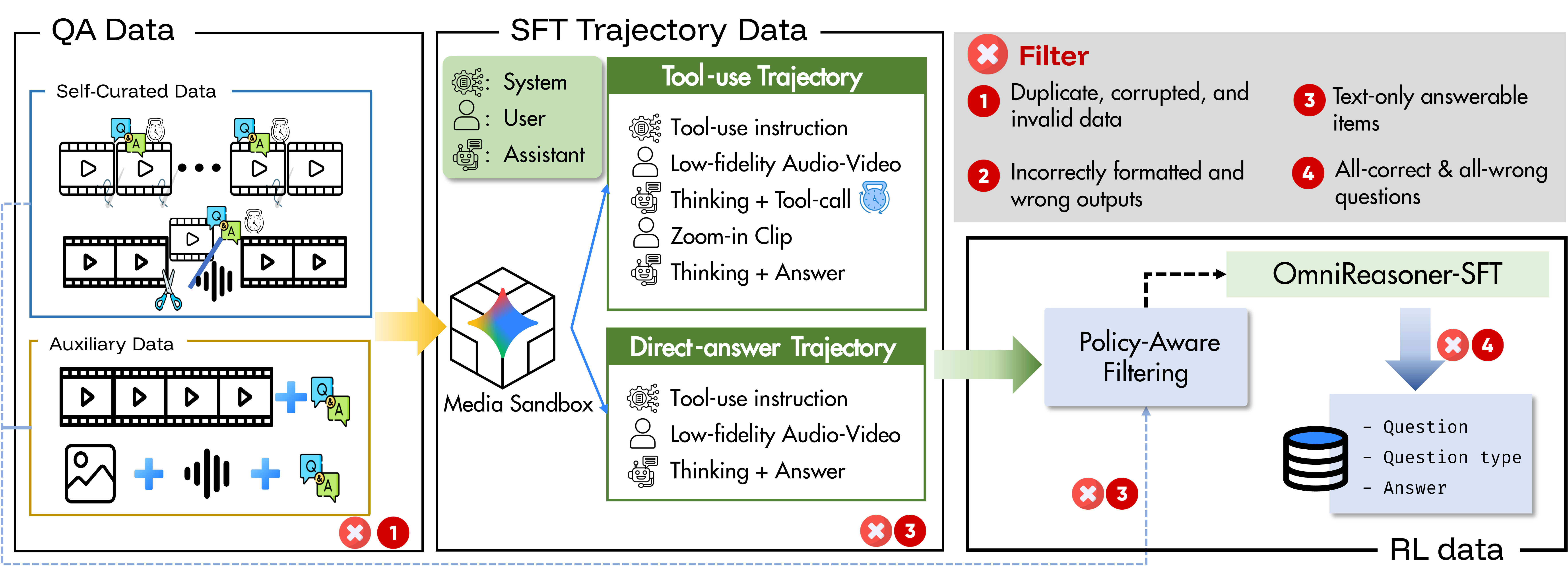}
    \caption{\textbf{Temporal Augmented Data Engine and post-training data.} Temporal editing creates long audio-video tasks with constructed evidence intervals. MediaSandbox converts these tasks into global-to-local tool-use trajectories, and filtering yields the SFT and difficulty-aware RL mixtures used to train OmniReasoner.}
    \label{fig:method_data_omnireasoner}
\end{figure*}

TimeAnchor makes the argument absolute with a deliberately simple construction: at the front of every two-second chunk on the global timeline, it prepends a plain-text marker \texttt{<t seconds>} (e.g.\ \texttt{<0 seconds>}, \texttt{<2 seconds>}, \dots) that binds the interleaved video and audio tokens of that chunk to wall-clock time on the original stream. The policy therefore reads and emits its zoom argument in absolute seconds rather than frame indices: a decision such as $\operatorname{zoom}([32,38])$ denotes seconds $32$--$38$ of the source, independent of how densely any stage samples the video. The media environment resolves this interval against the same absolute timeline to cut the higher-fidelity clip and its aligned audio, so the requested argument round-trips to the same moment before and after the tool call.

The mechanism adds no new special tokens, no architectural change, and no separate localization objective: the anchors are ordinary text tokens produced by the standard tokenizer. What TimeAnchor contributes is thus not a timestamp representation in itself, but the property that the zoom-in tool's temporal argument stays consistent across the fidelity boundary it must cross.

% \begin{figure*}[t]
%     \centering
%     \includegraphics[width=\textwidth]{figures/method-data-omnireasoner.pdf}
%     \caption{\textbf{Temporal Augmented Data Engine and post-training data.} Temporal editing creates long audio-video tasks with constructed evidence intervals. MediaSandbox converts these tasks into global-to-local tool-use trajectories, and filtering yields the SFT and difficulty-aware RL mixtures used to train OmniReasoner.}
%     \label{fig:method_data_omnireasoner}
% \end{figure*}

\subsection{Temporal Data Engine and Post-training Data}
\label{sec:data_construction}

Scalable open training data for long audio-video QA remains limited, especially when supervision must include both answers and evidence intervals. Directly asking strong MLLMs to synthesize QA pairs, or captioning audio/video before question generation, can be costly and prone to hallucinated or cross-modally inconsistent evidence in long streams. We therefore build a \textbf{Temporal Augmented Data Engine} that derives supervision from temporal editing operations. By composing and locally replacing video segments, the engine turns existing audio-video resources into long-form reasoning tasks whose evidence intervals are known by construction. The resulting tasks are curated with auxiliary omni data into SFT and RL post-training mixtures, as summarized in Figure~\ref{fig:method_data_omnireasoner}.

These constructed tasks provide the coupled signals OmniReasoner needs: answer supervision, interval supervision on the absolute-time timeline, and tool-use trajectory supervision linking global decisions to retrieved local evidence.

% \begin{figure*}[t]
%     \centering
%     \includegraphics[width=\textwidth]{figures/method-data-omnireasoner.pdf}
%     \caption{\textbf{Temporal Augmented Data Engine and post-training data.} Temporal editing creates long audio-video tasks with constructed evidence intervals. MediaSandbox converts these tasks into global-to-local tool-use trajectories, and filtering yields the SFT and difficulty-aware RL mixtures used to train OmniReasoner.}
%     \label{fig:method_data_omnireasoner}
% \end{figure*}

\textbf{Temporal augmented task construction.}
The engine instantiates temporal augmentation with two complementary task families, each designed to expose a different long audio-video reasoning behavior while keeping evidence intervals available by construction.

\textit{Multi-segment composition.} We stitch several semantically independent clips into a longer audio-video stream and attach each question to one source segment. The composed video records the start and end offset of every source clip, so the target segment offset directly becomes the evidence interval. This construction turns existing short or medium QA examples into long-context tasks: the model must scan through temporal distractors, identify the relevant segment, and answer from the local evidence rather than from the full stream uniformly.

\textit{Anomaly insertion.} We sample a temporal window in a long audio-video stream and replace the audio stream, the visual stream, or both with a short source clip. The model is trained to identify the anomaly type and localize the anomalous span. Because the replacement window is known during editing, it provides an interval label without manual span annotation. The three replacement modes create complementary cross-modal supervision.

Additional construction details and data statistics are reported in Appendix~\ref{sec:appendix_training_data}.

\textbf{Tool-use trajectory synthesis.}
Constructed intervals are converted into supervised trajectories through \texttt{MediaSandbox}, an FFmpeg-based audio-video sandbox we built to edit, downsample, and clip media on demand. It exposes the same global-to-local interface used at inference time: the model first receives a low-fidelity global observation, then either answers directly or requests an interval; when zoom is invoked, the sandbox materializes only the requested high-fidelity local clip. We synthesize two trajectory types. \emph{Tool-use trajectories} include a zoom-in call: the model observes the global view, requests an interval, receives the local clip, and answers from both observations. \emph{Direct-answer trajectories} skip the zoom step: the model answers directly from the global view when local evidence is unnecessary. We generate both types in two modes. In \emph{online synthesis}, the teacher model receives no reference answer or interval and generates the full trajectory from scratch; this mode is used for shorter videos and AVQA-R1 examples where the global observation is often sufficient. In \emph{hindsight synthesis}, the teacher receives the reference answer and interval as guidance but is explicitly instructed not to leak this supervision into the reasoning trace, producing trajectories that justify zoom decisions from observable cues; this mode is applied to longer, more challenging videos to improve sample utilization efficiency. Final stored trajectories are free of answer or reference-interval leakage in both modes.

\textbf{Curated SFT mixture.}
We curate 25,839 SFT examples from constructed temporal tasks and auxiliary omni data. Composition and anomaly insertion provide interval-by-construction supervision; FineVideo \citep{Farré2024FineVideo} adds open-ended long-video QA trajectories; AVQA-R1 \citep{xing2025echoinkr1exploringaudiovisualreasoning} preserves image-audio reasoning coverage; and CG-Bench \citep{queen2025cgbenchbenchmarkinglanguagemodel} adds public long-video QA diversity. Before training, we apply structural, leakage, answer, interval, and LLM-based grounding filters.

\textbf{Difficulty-aware RL data.}
For RL, we mine questions that produce useful reward variation under the SFT policy. We sample multiple rollouts per candidate and keep mixed-outcome examples, discarding cases solved by all rollouts or missed by all rollouts. We further apply policy-aware filtering to ensure the selected examples remain well-formed under policy exploration: this includes question-type consistency checks and answer-format validation. The final RL mixture contains 2,731 examples; detailed mixture statistics and filtering criteria are provided in Appendix~\ref{sec:appendix_training_data}.

Overall, the data recipe teaches OmniReasoner what to answer, when to retrieve higher-fidelity evidence, and where that evidence lies on the shared absolute-time timeline.

\subsection{Agentic Reinforcement Learning}
\label{sec:agentic_rl}

After supervised fine-tuning, the policy has already observed valid tool-use trajectories and can produce the required answer-or-interval format. We therefore use reinforcement learning to refine decision quality rather than to discover the interaction protocol from scratch. OmniReasoner uses no standalone reward for invoking zoom: a zoom action is favored only when it improves the final answer under the same format constraints.

We optimize the policy with Group Relative Policy Optimization (GRPO)~\citep{shao2024deepseekmathpushinglimitsmathematical}. For each prompt, we sample a group of trajectories from the old policy, where each trajectory either answers directly or invokes zoom before producing the final answer. Trajectory rewards are normalized within the group to compute relative advantages, and the policy is updated with the standard clipped GRPO objective. The reward is intentionally simple, $R=R_{\mathrm{acc}}+R_{\mathrm{fmt}}$: $R_{\mathrm{acc}}$ checks answer accuracy, and $R_{\mathrm{fmt}}$ checks whether the response follows the required protocol, including answer tags, interval tags when zoom is invoked, and valid second-based intervals. We do not assign a separate localization reward; the predicted interval is optimized only through its effect on final-answer correctness and trajectory validity.

\begin{table*}[ht]
\centering
\scriptsize
\setlength{\tabcolsep}{2.5pt}
\resizebox{\textwidth}{!}{%
\begin{tabular}{lcccccc}
\toprule
\multirow{2}{*}{\textbf{Method}} & \multicolumn{4}{c}{\textbf{Audio-Visual}} & \multicolumn{2}{c}{\textbf{General}} \\
\cmidrule(lr){2-5}\cmidrule(lr){6-7}
& \textbf{OmniVideoBench} & \textbf{LVOmniBench} & \textbf{Daily-Omni} & \textbf{WorldSense} & \textbf{VideoMME} & \textbf{VideoHolmes} \\
\midrule
\rowcolor{black!8}\multicolumn{7}{c}{\emph{Closed-source Models(Reference)}} \\
Gemini-2.0-Flash & 41.5 & 42.9 & 67.8 & 56.2 & 72.4 & 49.5 \\
Gemini-2.5-Pro & 58.9 & -- & 81.4 & 64.6 & 86.9 & 51.3 \\
\midrule
\rowcolor{black!8}\multicolumn{7}{c}{\emph{Open-source Models}} \\
VideoLLaMA2-7B~\cite{cheng2024videollama} & 29.2 & 27.2 & 35.2 & 25.4 & 62.4 & \underline{35.2} \\
MiniCPM-o-7B~\cite{cui2026minicpm} & 29.7 & -- & 53.1 & -- & \underline{63.9} & -- \\
VITA-1.5-7B~\cite{fu2025vita} & \underline{30.5} & -- & -- & 36.9 & 56.1 & -- \\
Ola-7B~\cite{liu2025ola} & -- & -- & 49.9 & -- & -- & -- \\
HumanOmniV2-7B & -- & -- & 58.5 & \textbf{47.1} & -- & -- \\
Qwen2.5-Omni-7B & 29.3 & \underline{32.0} & \underline{62.1} & 45.4 & \underline{64.3} & 24.4 \\
\rowcolor{blue!8}
\texttt{OmniReasoner} & \textbf{34.8}\gain{5.5} & \textbf{35.4}\gain{3.4} & \textbf{64.2}\gain{2.1} & \underline{46.7}\gain{1.3} & \textbf{65.4}\gain{1.1} & \textbf{40.0}\gain{15.6} \\
\bottomrule
\end{tabular}
}
\caption{Performance on audio-visual and general video reasoning benchmarks, including OmniVideoBench~\citep{li2026omnivideobench}, LVOmniBench~\citep{tao2026lvomnibench}, Daily-Omni~\citep{zhou2026dailyomni}, WorldSense~\cite{hong2026worldsenseevaluatingrealworldomnimodal}, VideoMME~\citep{fu2024video}, and VideoHolmes~\citep{cheng2025videoholmes}. The OmniReasoner row is highlighted.}
\label{tab:main_results}
\end{table*}

\section{Experiments}
\label{sec:experiments}

% \subsection{Experimental Setup}
% List the base omni model, datasets, sampling configurations, training protocol, and evaluation metrics.

\subsection{Implementation Details}
We use Qwen-Omni-7B as the base model. Both supervised fine-tuning and reinforcement learning are performed on 8 NVIDIA H100 GPUs with 80GB memory, requiring roughly 480 H100 GPU-hours for the two main post-training stages. During supervised fine-tuning, we train the model for 2 epochs with a batch size of 128, a learning rate of 5e-6, AdamW optimizer, and a cosine learning-rate scheduler, using the \texttt{LMMS-Engine} framework. For reinforcement learning, we implement an omni-modal tool-use RL framework based on \texttt{TRL}~\cite{vonwerra2020trl}, as existing RL frameworks such as \texttt{veRL}~\cite{sheng2024hybridflow} do not directly support omni-modal models with audio-conditioned inputs and tool-use interactions. We train with a batch size of 64, sample 8 rollouts per prompt, set the KL coefficient to 0, use a maximum response length of 8192 tokens, and set the learning rate to 1e-6. Evaluation is conducted mainly with \texttt{LMMS-Eval}~\cite{Zhang_2025}.

\subsection{Main Results}

Table~\ref{tab:main_results} compares OmniReasoner on audio-visual benchmarks and general video reasoning benchmarks. Compared with the Qwen2.5-Omni-7B base model, OmniReasoner improves OmniVideoBench from 29.3 to 34.8, LVOmniBench from 32.0 to 35.4, Daily-Omni from 62.1 to 64.2, WorldSense from 45.4 to 46.7, VideoMME from 64.3 to 65.4, and VideoHolmes from 24.4 to 40.0. These gains indicate that tool-use post-training improves both audio-visual reasoning and broader video reasoning, with particularly large improvements on benchmarks where the answer depends on locating sparse evidence before detailed reasoning. The VideoHolmes gain is partly due to the baseline using only 32 frames per video, while OmniReasoner retrieves more frames via zoom-in.

The gains on Daily-Omni and WorldSense are more moderate, with improvements from 62.1 to 64.2 and from 45.4 to 46.7. This is expected: Daily-Omni and WorldSense mostly involve short to medium audio-video clips, while OmniVideoBench and especially LVOmniBench place stronger pressure on long-context temporal search and sparse evidence localization. Importantly, OmniReasoner still improves over the base model on both benchmarks, suggesting that tool-use post-training for long audio-video reasoning does not come at the cost of general audio-visual competence.

\begin{table}[h]
\centering
\scriptsize
\setlength{\tabcolsep}{1pt}
\resizebox{\columnwidth}{!}{%
\begin{tabular}{@{}l@{\hspace{3pt}}c@{\hspace{3pt}}c@{\hspace{5pt}}c@{\hspace{8pt}}c@{\hspace{5pt}}c@{\hspace{3pt}}c@{}}
\toprule
\multirow{2}{*}{\textbf{Method}} & \multicolumn{3}{c}{\textbf{Audio Type}} & \multicolumn{3}{c}{\textbf{Video Duration}} \\
\cmidrule(lr){2-4}\cmidrule(lr){5-7}
& Music & Sound & Speech & 0-5 min & 5-10 min & 10-30 min \\
\midrule
\rowcolor{black!8}\multicolumn{7}{c}{\emph{Closed-source Models}} \\
Gemini-2.0-Flash & 29.7 & 40.3 & 43.2 & 45.3 & 41.1 & 34.9 \\
Gemini-2.5-Pro & 38.5 & 57.7 & 61.7 & 62.2 & 55.0 & 55.9 \\
Gemini-3-Pro & 56.2 & 54.1 & 55.7 & 58.0 & 52.9 & 52.5 \\
\midrule
\rowcolor{black!8}\multicolumn{7}{c}{\emph{Open-source Models}} \\
VideoLLaMA2-7B & 26.4 & \underline{30.7} & 29.3 & 29.5 & 29.6 & 28.3 \\
MiniCPM-o-7B & \underline{27.5} & 28.6 & 30.2 & 29.5 & \textbf{34.5} & 26.2 \\
HumanOmniV2-7B & 20.9 & \textbf{31.1} & \underline{31.6} & 31.8 & 29.6 & \underline{29.3} \\
Qwen2.5-Omni-7B & 23.1 & 25.3 & 30.7 & \underline{32.2} & 25.3 & 26.7 \\
\rowcolor{blue!8}
\texttt{OmniReasoner} & \textbf{34.1}\gain{11.0} & 29.3\gain{4.0} & \textbf{36.0}\gain{5.3} & \textbf{35.4}\gain{3.2} & \underline{31.9}\gain{6.6} & \textbf{36.6}\gain{9.9} \\
\bottomrule
\end{tabular}
}
\caption{Accuracy comparison on OmniVideoBench by audio type and video duration. All duration buckets are reported in minutes.}
\label{tab:duration_results}
\end{table}

\begin{figure}[h]
\centering
\includegraphics[width=\columnwidth]{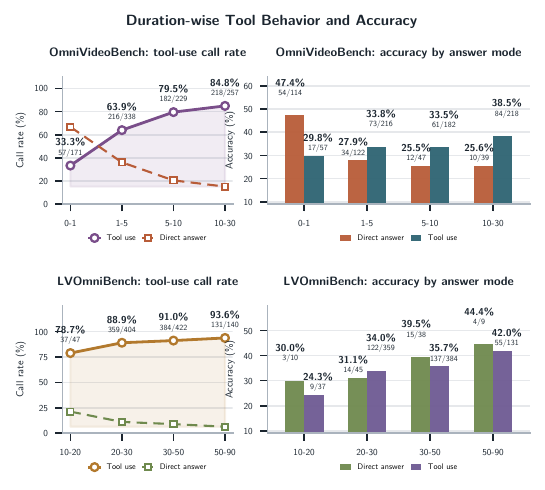}
\caption{Duration-wise tool behavior on LVOmniBench.}
\label{fig:duration_tool_behavior}
\end{figure}

Table~\ref{tab:duration_results} further compares OmniVideoBench performance by audio type and video duration. The duration split shows a clear trend: compared with Qwen2.5-Omni-7B, OmniReasoner's gains increase from 3.2 points on 0--5 min videos to 6.6 points on 5--10 min videos and 9.9 points on 10--30 min videos. This suggests that adaptive zoom-in becomes more useful as the input grows longer and sparse evidence is harder to preserve under uniform processing. The audio-type split shows consistent gains across music, sound, and speech.

% \subsection{Ablations}
% Isolate the contribution of tool-use post-training, the temporal augmented data engine, and TimeAnchor.

% \subsection{Analysis}
% Analyze when the model chooses to zoom in, how often the retrieved interval overlaps with useful evidence, and failure cases where temporal grounding or audio-video coordination breaks down.

\subsection{Ablation Studies}

\begin{table}[h]
\centering
\small
\setlength{\tabcolsep}{3pt}
\resizebox{\columnwidth}{!}{%
\begin{tabular}{lccc}
\toprule
\textbf{Model} & \textbf{OmniVideoBench} & \textbf{LVOmniBench} & \textbf{WorldSense} \\
\midrule
\multicolumn{4}{c}{\emph{Data Recipe}} \\
\midrule
w/o self curated data & 31.0\abldrop{3.8} & 29.8\abldrop{5.6} & 44.1\abldrop{2.6} \\
w/o tool-use data & 32.8\abldrop{2.0} & 33.3\abldrop{2.1} & 44.5\abldrop{2.2} \\
\midrule
\multicolumn{4}{c}{\emph{Input Configuration}} \\
\midrule
w/o audio input & 30.9\abldrop{3.9} & 31.3\abldrop{4.1} & 38.4\abldrop{8.3} \\
w/o tool-return & 32.6\abldrop{2.2} & 33.9\abldrop{1.5} & 44.7\abldrop{2.0} \\
\midrule
\multicolumn{4}{c}{\emph{Training Stage}} \\
\midrule
sft & 32.9\abldrop{1.9} & 34.2\abldrop{1.2} & 45.1\abldrop{1.6}  \\
rl-only & 27.6\abldrop{7.2} & 26.7\abldrop{8.7} & 38.4\abldrop{8.3} \\
\rowcolor{blue!8}
Full & 34.8 & 35.4 & 46.7 \\
\bottomrule
\end{tabular}
}
\caption{Ablation study across data recipe, input configuration, and training stage.}
\label{tab:recipe_input_training_ablation}
\end{table}

\textbf{Effect of data recipe.} Table~\ref{tab:recipe_input_training_ablation} first evaluates the contribution of the data recipe. The ``w/o self curated data'' variant removes our constructed long audio-video QA data, including both multi-segment composition and anomaly insertion tasks, and trains only on the remaining data. This variant drops substantially on OmniVideoBench and LVOmniBench, showing that the constructed data is important for learning long audio-video reasoning and temporal grounding. The ``w/o tool-use data'' variant keeps only direct-answer samples from the data mixture and removes tool-use trajectories by rewriting tool-related instructions into ordinary chain-of-thought supervision. This variant corresponds to a standard CoT-style setting in which the model thinks and then directly answers, rather than actively acquiring local evidence from the long audio-video input. Its lower performance highlights the advantage of our thinking-with-long-audio-video paradigm over ordinary CoT for long-context audio-video reasoning. The self-curated-data ablation performs even worse than the direct-answer variant, likely because the remaining data neither provides enough long audio-video collaborative reasoning examples nor teaches the tool-use paradigm reliably.

\textbf{Effect of input configuration.} We further ablate the input configuration at inference time. Removing the audio track causes clear drops across all benchmarks, especially on WorldSense, showing that OmniReasoner does not solve these tasks from visual cues alone and still relies on audio-visual evidence. Removing the tool-returned zoom-in clip also hurts performance, which indicates that the model benefits from actively acquired local evidence rather than relying only on the first-stage global input.

\textbf{Effect of training stage.} We finally ablate the post-training stages. The SFT-only model corresponds to the cold-start checkpoint after supervised tool-use training, while the RL-only variant applies reinforcement learning directly to Qwen2.5-Omni-7B. Direct tool-use RL is difficult for the base omni model: without a supervised warm-up, the model is reluctant to call the tool, so we use a fixed two-stage template that first forces a global observation and zoom-in step and then asks the model to answer in the second stage. Even under this constrained protocol, RL-only drops sharply across all benchmarks. This suggests that cold-start SFT is necessary for establishing the tool-use format, the basic zoom-in policy, and stable answer behavior before RL optimization. Without this foundation, the policy drifts severely during RL, and after more training steps the model can even lose the required output format.

\begin{table*}[t]
\centering
\small
\setlength{\tabcolsep}{7pt}
\resizebox{0.82\textwidth}{!}{%
\begin{tabular}{lcccccc}
\toprule
\multirow{2}{*}{\textbf{Model}} & \multirow{2}{*}{\textbf{OmniVideoBench}} & \multirow{2}{*}{\textbf{LVOmniBench}} & \multicolumn{4}{c}{\textbf{Charades-STA}} \\
\cmidrule(lr){4-7}
& & & \textbf{IoU@0.3} & \textbf{IoU@0.5} & \textbf{IoU@0.7} & \textbf{mIoU} \\
\midrule
Qwen2.5-Omni-7B & 29.3 & 32.0 & 53.3 & 31.7 & 13.4 & 33.7 \\
OmniReasoner w/o TimeAnchor & 32.3\gain{3.0} & 32.8\gain{0.8} & 58.8\gain{5.5} & 37.5\gain{5.8} & 16.2\gain{2.8} & 37.9\gain{4.2} \\
\rowcolor{blue!8}
\texttt{OmniReasoner} & \textbf{34.8}\gain{5.5} & \textbf{35.4}\gain{3.4} & \textbf{64.9}\gain{11.6} & \textbf{41.1}\gain{9.4} & \textbf{17.7}\gain{4.3} & \textbf{41.3}\gain{7.6} \\
\bottomrule
\end{tabular}
}
\caption{Ablation of TimeAnchor on OmniVideoBench, LVOmniBench, and Charades-STA~\cite{gao2017tall}. All scores are reported as percentages.}
\label{tab:timeanchor_ablation}
\end{table*}

\textbf{Effect of TimeAnchor.} Table~\ref{tab:timeanchor_ablation} isolates the contribution of TimeAnchor. Besides OmniVideoBench and LVOmniBench, we include Charades-STA because it directly evaluates video grounding: the model must localize the temporal segment that matches a natural-language query. Removing TimeAnchor at inference time still improves over the Qwen2.5-Omni-7B baseline, indicating that our training data and recipe already improve the model's temporal grounding ability. However, adding TimeAnchor yields substantially stronger grounding performance, raising Charades-STA from 58.8 to 64.9 on IoU@0.3, from 37.5 to 41.1 on IoU@0.5, and from 37.9 to 41.3 on mIoU. The gains are consistent with its design: second-based anchors give the model a stable timeline when moving between sparse global observations and dense local clips, making selected intervals easier to align with the original video. The same effect also transfers back to answer accuracy, improving OmniVideoBench from 32.3 to 34.8 and LVOmniBench from 32.8 to 35.4 over the variant without TimeAnchor.

\subsection{Analysis}

\textbf{Tool-use behavior by duration.} Figure~\ref{fig:duration_tool_behavior} analyzes how OmniReasoner allocates tool calls as input duration changes. The left panels show a clear routing pattern: longer videos are more likely to trigger the zoom-in tool. On OmniVideoBench, the tool-use call rate increases from 33.3\% in the 0--1 min bucket to 84.8\% in the 10--30 min bucket; on LVOmniBench, it further rises from 78.7\% for 10--20 min videos to 93.6\% for 50--90 min videos. This behavior is consistent with the tool-use motivation behind OmniReasoner: as videos become longer, decisive evidence becomes harder to preserve in the global view, so the model more often calls the zoom-in tool. The right panels compare accuracy under direct-answer and tool-use answer modes. Tool-use accuracy tends to improve as duration increases, suggesting that adaptive zoom-in is especially beneficial for long videos. In some duration buckets, direct answers can still show higher accuracy than tool-use answers; this does not imply that direct answering is generally better. Rather, OmniReasoner tends to answer directly on simpler or more certain examples, while reserving tool use for cases that require additional evidence.

% \begin{figure*}[h]
%     \centering
%     \includegraphics[width=\textwidth]{figures/attn-rollout.pdf}
%     \caption{Attention rollout visualization for OmniReasoner. The examples show how the model predicts a TimeAnchor-grounded interval and uses the returned zoom-in clip to answer, with rollout maps projected onto the video timeline.}
%     \label{fig:attn_rollout}
% \end{figure*}

\textbf{Are interleaved audio-video cues causally used?} Recent work questions whether interleaved visual cues inserted into chain-of-thought are faithfully used for prediction, rather than merely appearing in the reasoning trace~\cite{liu2025faithfulnessvisualthinkingmeasurement}. To test this issue in our setting, we conduct a direct intervention by removing the tool-returned zoom-in clip at inference time. As shown by the ``w/o tool-return'' row in Table~\ref{tab:recipe_input_training_ablation}, performance drops after this removal, indicating that the returned audio-video evidence is not decorative. We further provide an internal-view analysis in Figure~\ref{fig:attn_rollout} using attention rollout~\cite{abnar2020quantifying,xu2026flowprune}. We teacher-force the full stage-2 response, including both CoT and final-answer tokens, and aggregate their causal attention rollout over the last transformer layers. Media-token attributions are then grouped into Qwen2.5-Omni 2-second chunks and visualized for the full-video context and the zoom-in clip on the same temporal axis. The rollout concentrates strongly on the stage-2 zoom-in clip, suggesting that OmniReasoner attends to the interleaved audio-video cues when reasoning toward the final answer.

\section{Conclusion}
\label{sec:conclusion}

We presented OmniReasoner, a tool-use post-training framework for long audio-video reasoning. OmniReasoner learns to call a zoom-in tool adaptively, supported by a Temporal Augmented Data Engine and a curated SFT/RL mixture for learning when to call the tool and how to answer from retrieved evidence. TimeAnchor keeps the tool's temporal argument valid across changing sampling granularities, enabling omni models to allocate high-fidelity perception to evidence-bearing moments.

\section*{Limitations}
\label{sec:limitations}

\textbf{Infrastructure and model capacity.} OmniReasoner focuses on a two-step tool-use process in part because the infrastructure for multi-turn agentic RL on omni-modal models remains underdeveloped: existing RL frameworks such as veRL and OpenRLHF are designed for text-only or vision-language models and do not natively support audio-conditioned inputs and repeated audio-video tool returns. We implemented our own omni-modal tool-use RL framework on top of TRL, but this setup is substantially less mature than text or vision counterparts. Additionally, we use Qwen2.5-Omni-7B, a relatively small model with a 32K context window. This base model was not pretrained with significant agentic or tool-use data, so direct RL or lightweight SFT without a strong supervised warm-up leads to very poor tool-use behavior: the policy either refuses to call the tool or produces malformed intervals and answers. Scaling to longer multi-turn reasoning therefore requires both stronger omni-modal RL infrastructure and larger, longer-context base models pretrained with tool-aware data. We did not experiment with Qwen-Omni-3, a larger and more recent model, due to limited compute resources and our team's relative inexperience with large-scale infrastructure engineering. Future work with such models and improved training frameworks could extend this paradigm beyond two-step zoom-in.

\textbf{Tool scope.} OmniReasoner implements only a single perception-focused tool---the zoom-in tool for retrieving higher-fidelity audio-video clips. We do not explore complementary tool types such as web search for retrieving external knowledge, code execution for performing symbolic reasoning over extracted information, or structured query tools for accessing databases or knowledge graphs. This narrow tool scope limits OmniReasoner's ability to handle tasks that require combining perceptual evidence with retrieved facts, computational steps, or external domain knowledge. Expanding the tool repertoire and training the model to coordinate multiple tools remain important directions for more general omni-modal reasoning agents.

\section*{Ethics Statement}

OmniReasoner is intended for research on long audio-video reasoning. The training recipe uses public or curated audio-video resources and automatically generated tool-use trajectories; dataset licensing, privacy-sensitive content, and benchmark-specific usage constraints should be checked before public release. Public audio-video sources may contain identifiable people, voices, scene text, or offensive content inherited from the original datasets. The model may still produce incorrect temporal grounding or hallucinated audio-visual evidence, so downstream use should include human verification in high-stakes settings.
We use external datasets, benchmarks, models, and software frameworks under their respective licenses and terms of use. Any released derivative artifacts will follow the access and redistribution constraints of the original sources, and benchmark data will not be redistributed when the original terms prohibit redistribution.

% Acknowledgments will be added after acceptance.

\bibliography{main}

%%%%%%%%%%%%%%%%%%%%%%%%%%%%%%%%%%%%%%%%%%%%%%%%%%%%%%%%%%%%

\clearpage
\appendix

% \section{Appendix Overview}
% Appendix sections can later hold implementation details, prompt templates, additional examples from the temporal augmented data engine, extended quantitative results, and failure case analysis.

\begin{figure*}[t]
    \centering
    \includegraphics[width=\textwidth]{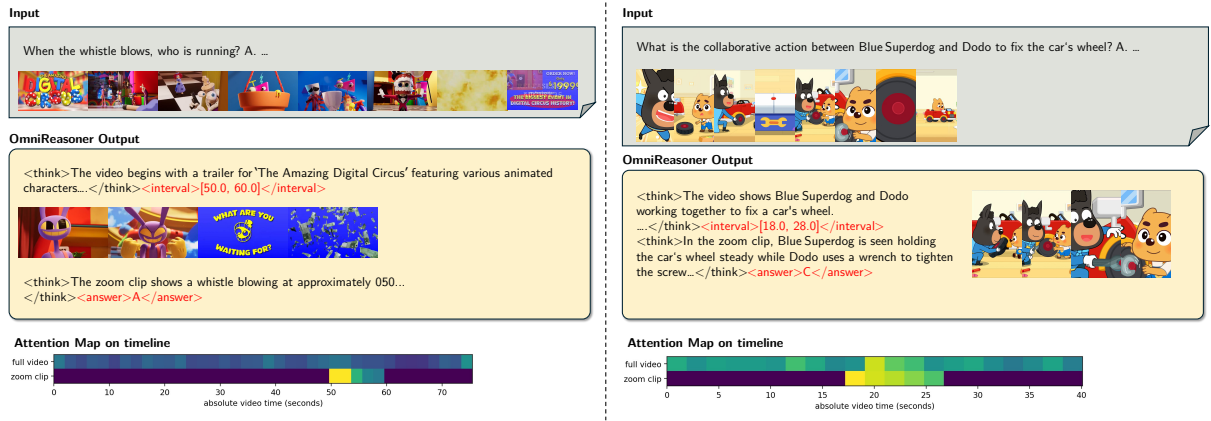}
    \caption{Attention rollout visualization for OmniReasoner. The examples show how the model predicts a TimeAnchor-grounded interval and uses the returned zoom-in clip to answer, with rollout maps projected onto the video timeline.}
    \label{fig:attn_rollout}
\end{figure*}

\section{Temporal Augmented Data Engine and Training Data}
\label{sec:appendix_training_data}

\subsection{Temporal Task Construction}

The Temporal Augmented Data Engine derives interval supervision from editing operations rather than manual span annotation. For multi-segment composition, we build a candidate pool from FineVideo videos with available QA annotations and valid local media. We sample a target duration, greedily stitch multiple semantically independent clips into a longer stream, and attach each question to one source segment. The composed video stores the start and end offset of every source clip, so the target segment offset becomes the evidence interval. This route places existing QA pairs into a long temporal context and trains the model to find the relevant segment before answering.

For anomaly insertion, we use long CG-Bench videos as main videos and short LLaVA-Video-178K clips as replacement sources. The engine samples a temporal window and replaces only the audio stream, only the visual stream, or both modalities. The output keeps the main-video timeline, while the replacement window gives the anomaly interval by construction. The three replacement types produce complementary supervision: audio-only replacements require detecting sound-visual inconsistency, video-only replacements require detecting visual inconsistency against the original audio, and both-modality replacements require detecting a local event that conflicts with the surrounding context. For audio-related anomaly samples, we keep only cases whose anomaly window overlaps the observable audio range in the global stage, so the first scan contains evidence the model can learn from.

\subsection{Trajectory Synthesis and Filtering}

MediaSandbox turns constructed intervals into the same global-to-local protocol used by OmniReasoner. In an interval trajectory, the model first receives a low-fidelity global observation and the question, outputs a TimeAnchor-grounded interval, receives the zoomed local clip from that interval, and then produces the final answer. In a direct-answer trajectory, the model answers from the global observation when local evidence is unnecessary. The sandbox materializes only the global preview and requested local clip, keeping synthesis and training within a bounded token budget.

We use \texttt{Gemini-3-Flash-Preview}~\cite{team2023gemini} to synthesize tool-use trajectories. Online synthesis does not expose the final answer during generation and keeps only trajectories whose final answers match the ground-truth label. Hindsight synthesis uses the final answer and reference interval as generation-time guidance for harder long-context cases, while the stored trajectory must justify zoom decisions from observable cues and answer from the returned clip. We filter generated data with structural rules for message topology, tags, answer format, and interval bounds; leakage checks for explicit answer or generation-metadata leakage; task-specific answer verification; and an LLM~\cite{deepseekai2025deepseekv32pushingfrontieropen} judge for evidence grounding, reasoning quality, hallucination risk, and consistency between global and local stages.

\subsection{Post-training Mixtures}

\begin{figure*}[t]
    \centering
    \includegraphics[width=\textwidth]{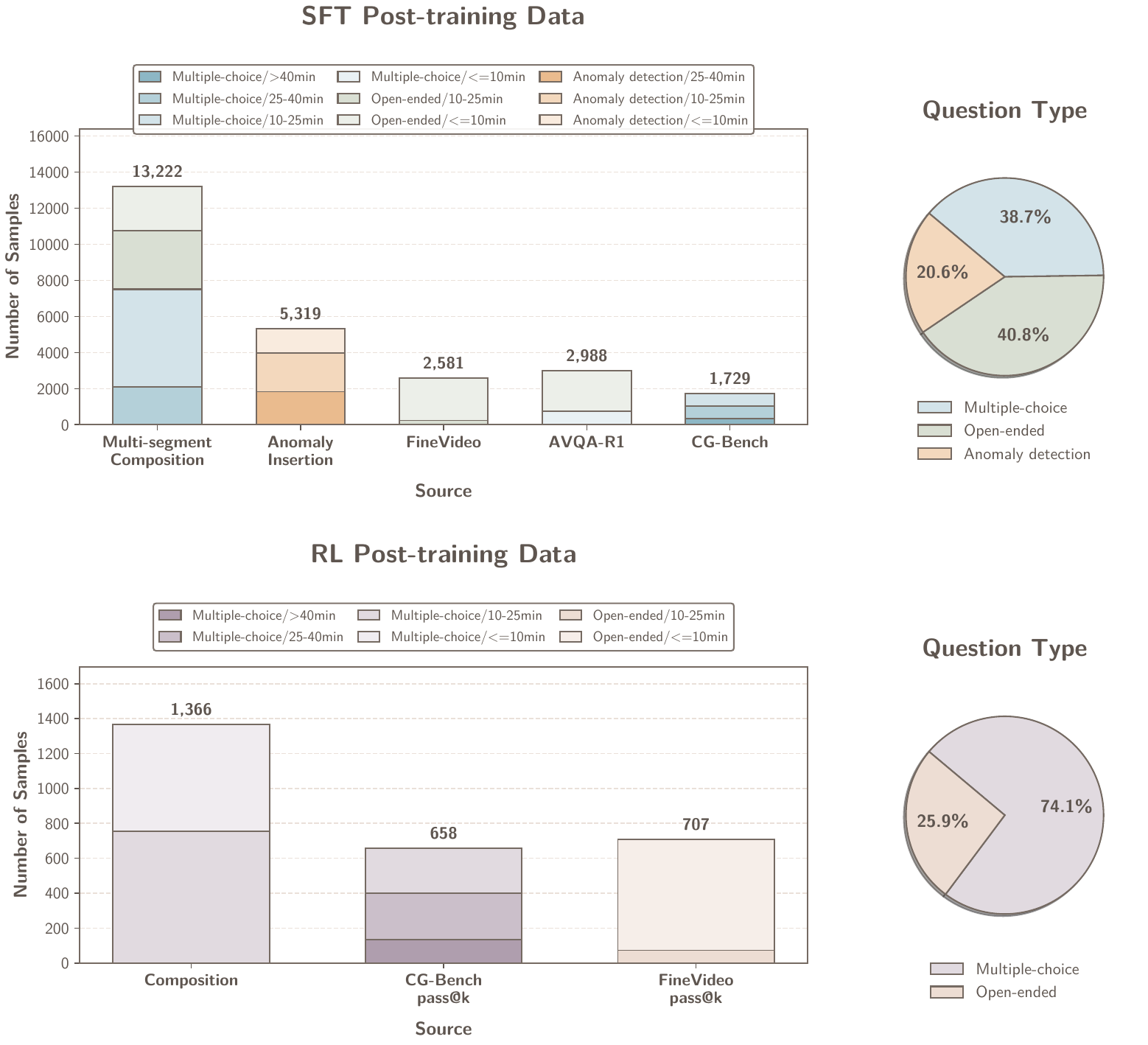}
    \caption{Post-training data used by OmniReasoner. The SFT mixture provides answer, interval, and tool-use trajectory supervision. The RL mixture contains examples selected to produce non-degenerate group-relative rewards under the SFT policy. Bars show per-source counts split by question type and duration, while pie charts show the overall question-type composition for each split.}
    \label{fig:post_training_data_stats}
\end{figure*}

Figure~\ref{fig:post_training_data_stats} summarizes the post-training data used by OmniReasoner. The SFT mixture contains 25,839 examples and combines interval-by-construction temporal tasks with auxiliary omni-modal reasoning data. Multi-segment composition contributes 13,222 examples, anomaly insertion contributes 5,319 examples, FineVideo~\cite{Farré2024FineVideo} contributes 2,581 open-ended tool-use trajectories, AVQA-R1~\cite{xing2025echoinkr1exploringaudiovisualreasoning} contributes 2,988 image-audio examples, and CG-Bench~\cite{queen2025cgbenchbenchmarkinglanguagemodel} contributes 1,729 long-video QA examples. These sources provide complementary supervision: constructed temporal tasks expose the model to known evidence intervals, FineVideo adds real open-ended long-video tool-use trajectories, AVQA-R1 preserves image-audio coverage, and CG-Bench broadens public long-video QA diversity.

The SFT mixture contains 21,032 interval trajectories, 1,819 direct-answer trajectories, and 2,988 AVQA-R1 image-audio examples. For RL, we mine examples that produce non-degenerate group-relative rewards under the SFT policy. Composition candidates are converted to MCQ format, filtered for text-only guessability, and then sampled with eight SFT-policy rollouts; we keep examples with at least one but not all rollouts correct. CG-Bench and FineVideo RL examples are mined with the same pass@k-but-not-pass@1 criterion. The final RL mixture contains 2,731 examples: 1,366 composition examples, 658 CG-Bench examples, and 707 FineVideo examples. The stacked bars in Figure~\ref{fig:post_training_data_stats} show the joint distribution of question type and input duration for each source. Multiple-choice, open-ended, and anomaly-detection examples are shown with distinct color families, and longer-duration bins use darker shades within the same question type. AVQA-R1 is treated as image-audio data and assigned to the shortest duration bin because it does not provide long-video duration metadata.

\clearpage
\section{Prompt}

\subsection{Inference-time prompts}

The base protocol below governs every inference call; for multiple-choice examples, the task adapter additionally appends \texttt{MULTIPLE\_CHOICE\_CONSTRAINT} to the first-stage user question only (open-ended examples do not receive this constraint), while anomaly-detection examples instead receive \texttt{ANOMALY\_ANSWER\_CONSTRAINT}, which fixes the answer to a typed anomaly label and a second-based interval.

\begin{strip}
\begin{promptbox}{promptblue}{SYSTEM\_PROMPT}
\begin{Verbatim}[fontsize=\scriptsize,breaklines=true,breakanywhere=true]
You are an audio-visual assistant.
Output tags only, with no text outside tags.
Use exactly one:
  <think>...</think><interval>[start, end]</interval>
  <think>...</think><answer>...</answer>
Never output both <interval> and <answer>.
All time values are in seconds on the same timeline.
\end{Verbatim}
\end{promptbox}

\begin{promptbox}{promptteal}{USER\_PROMPT}
\begin{Verbatim}[fontsize=\scriptsize,breaklines=true,breakanywhere=true]
If current evidence is sufficient, output <answer>...</answer>.
Otherwise, request one zoom interval with <interval>[start, end]</interval>.
\end{Verbatim}
\end{promptbox}

\begin{promptbox}{promptorange}{RETURN\_ZOOM\_IN\_USER\_PROMPT}
\begin{Verbatim}[fontsize=\scriptsize,breaklines=true,breakanywhere=true]
Zoom-in video clip [start, end] seconds:
Use the provided zoom-in clip (and prior context) to output final <answer>...</answer>.
Do not output <interval> in this turn.

<zoom-in audio-video clip>
\end{Verbatim}
\end{promptbox}

\begin{promptbox}{promptviolet}{MULTIPLE\_CHOICE\_CONSTRAINT}
\begin{Verbatim}[fontsize=\scriptsize,breaklines=true,breakanywhere=true]
Answer with exactly one option letter from the provided choices.
\end{Verbatim}
\end{promptbox}

\begin{promptbox}{promptviolet}{ANOMALY\_ANSWER\_CONSTRAINT}
\begin{Verbatim}[fontsize=\scriptsize,breaklines=true,breakanywhere=true]
If <answer> is used, format exactly:
type=<audio_only|video_only|both>; interval=[start, end]
Do not add extra words inside <answer>.
\end{Verbatim}
\end{promptbox}

\subsection{Trajectory synthesis: hindsight teacher prompts}
\label{sec:appendix_teacher_prompts}

Interval-trajectory SFT data (\S\ref{sec:appendix_training_data}) is synthesized with \texttt{Gemini-3-Flash-Preview} acting as a hindsight teacher: the teacher is given the reference answer and interval as supervision, but its system prompt explicitly forbids leaking that supervision into the produced reasoning, so the resulting trajectory reads as first-pass, evidence-grounded reasoning rather than answer-conditioned rationalization. The same global-skim / zoom-verification contract is used for multi-segment composition and for anomaly detection, with the four system prompts below differing only in the task-specific fields and vocabulary each phase must produce.

\begin{promptbox}{promptmagenta}{COMPOSITION\_TEACHER\_STAGE1}
\begin{Verbatim}[fontsize=\scriptsize,breaklines=true,breakanywhere=true]
You are a hindsight teacher for long-video multiple-choice reasoning on a MULTI-SEGMENT composed video.
PHASE-1: global skim and zoom planning from a low-quality full-video overview.

Output exactly one JSON object with keys: think_stage1, decision, interval, answer.

Hard constraints:
- think_stage1: 4-7 sentences. Must include: (1) at least one concrete time anchor tied to observations (not copied blindly from any "reference interval" wording); (2) reasoning that connects the provided segment description to what you see/hear while scanning; (3) why zoom is or is not needed.
- decision must be 'interval' or 'answer'. If 'interval', it must be [start, end] in seconds, end > start, and should land on the segment matching the description.
- If decision='answer', answer must be exactly one valid option label.
- Do not output markdown/code fences or any extra text outside JSON.
- Do not mention supervision, reference answer, ground truth, or that you were given hidden labels.
- Do NOT restate the reference interval as if it were user-visible information; simulate first-pass viewing.
\end{Verbatim}
\end{promptbox}

\begin{promptbox}{promptmagenta}{COMPOSITION\_TEACHER\_STAGE2}
\begin{Verbatim}[fontsize=\scriptsize,breaklines=true,breakanywhere=true]
You are a hindsight teacher for long-video multiple-choice reasoning on a MULTI-SEGMENT composed video.
PHASE-2: zoom verification and final decision using the zoom clip plus prior turn context.

Output exactly one JSON object with keys: think_stage2, answer.

Hard constraints:
- think_stage2: 4-7 sentences with evidence from the zoom clip, integration/revision vs the global view, and final rationale. Address whether the segment description matches the zoomed evidence.
- You may revise or overturn the Stage1 hypothesis if zoom evidence contradicts it.
- answer must be exactly one valid option label.
- Do not output markdown/code fences or any extra text outside JSON.
- Do not mention supervision, reference answer, ground truth, or hidden labels.
\end{Verbatim}
\end{promptbox}

\begin{promptbox}{promptmagenta}{ANOMALY\_TEACHER\_STAGE1}
\begin{Verbatim}[fontsize=\scriptsize,breaklines=true,breakanywhere=true]
You are a hindsight teacher for long-video ANOMALY DETECTION (audio/visual/both).
PHASE-1: fast global overview (low FPS / low resolution). Plan whether to zoom into a time window for verification.

Output exactly one JSON object with keys: think_stage1, decision, interval, answer_direct

- think_stage1: 4-8 sentences. Must sound like a first-time viewer: observations, uncertainty, self-check ("Could this be intentional editing?"), and why zoom helps or not. Do not reveal that you were given hidden labels.
- decision: "interval" (need zoom-in clip to verify) or "answer" (global view is sufficient).
- interval: [start_s, end_s] in seconds, required when decision is "interval" (end > start). Refine the suggested window from observable cues; do not paste supervision as your only justification.
- answer_direct: only when decision is "answer", a string EXACTLY of the form type=<audio_only|video_only|both>; interval=[start, end]; otherwise null.

Hard constraints:
- JSON only, no markdown fences, no extra keys.
- Do not mention "ground truth", "supervision", or "reference interval" in natural language.
- think_stage1 must NOT quote the reference interval numbers alone as proof; tie times to observable cues.
\end{Verbatim}
\end{promptbox}

\begin{promptbox}{promptmagenta}{ANOMALY\_TEACHER\_STAGE2}
\begin{Verbatim}[fontsize=\scriptsize,breaklines=true,breakanywhere=true]
You are a hindsight teacher for long-video ANOMALY DETECTION.
PHASE-2: examine the zoom-in clip (higher detail) and output the FINAL structured detection result.

Output exactly one JSON object with keys: think_stage2, final_answer

- think_stage2: 5-10 sentences. Compare zoomed evidence to the global impression; justify anomaly type (audio-only / video-only / both) and the tight time bounds.
- final_answer: a single structured string EXACTLY of the form type=<audio_only|video_only|both>; interval=[start, end], in seconds.

Hard constraints:
- JSON only, no markdown fences.
- Do not mention supervision or hidden labels.
- Base bounds on zoom clip evidence; avoid claiming arbitrary precision without cues.
\end{Verbatim}
\end{promptbox}

\subsection{Trajectory quality judge}
\label{sec:appendix_judge_prompt}

Synthesized trajectories are additionally screened by an LLM judge (\S\ref{sec:appendix_training_data}) that scores reasoning quality, answer correctness, leakage risk, and audio-visual grounding quality on a 1--10 scale; only trajectories above the acceptance threshold enter the SFT mixture. The composition judge is shown below; the open-ended and anomaly-detection judges share the same rubric structure but swap in task-appropriate fields (\texttt{detection\_accuracy} and \texttt{temporal\_precision} in place of \texttt{answer\_correctness} for anomaly detection).

\begin{promptbox}{promptgray}{TRAJECTORY\_JUDGE\_PROMPT (composition)}
\begin{Verbatim}[fontsize=\scriptsize,breaklines=true,breakanywhere=true]
You are an expert evaluator for long-video multi-segment MCQ reasoning trajectories.

## Task
Evaluate the quality of a two-stage (or single-stage) Chain-of-Thought (CoT) trajectory for a multiple-choice question on a multi-segment composed long video.

## Input Fields
question, options, correct_answer, pred_answer, think_stage1, think_stage2 (if applicable), duration, gt_intervals, zoomin_interval.

## Scoring Dimensions (each 1-10)
1. reasoning_quality: Does the CoT demonstrate genuine temporal reasoning? Does it tie observations to specific time anchors and use the segment description to narrow down the correct segment? Deduct for vague, repetitive, or tautological reasoning.
2. answer_correctness: Strict gate -- 10 if exact match to correct_answer, 1 if not. Do not use middle scores.
3. leakage_risk: 10 = no leakage signs; 1 = obvious leakage (quotes supervision/GT verbatim, uses "reference"/"ground truth" language, or states the answer without evidence).
4. grounding_quality: Does the reasoning cite specific visual/audio cues rather than just parroting the segment description?

## Output
Return ONLY a JSON object (no markdown fences):
{"reasoning_quality": N, "answer_correctness": N, "leakage_risk": N, "grounding_quality": N, "overall_score": N, "decision": "accept|borderline|reject", "brief_rationale": "..."}

overall_score = reasoning_quality*0.3 + answer_correctness*0.3 + leakage_risk*0.2 + grounding_quality*0.2
\end{Verbatim}
\end{promptbox}
\end{strip}

\subsection{MCQ design and text-only guessability filter}
\label{sec:appendix_mcq_prompts}

Composition candidates (\S\ref{sec:appendix_training_data}) are first converted from open QA into 4-option MCQ format by an LLM designer explicitly prompted against option-length and structural shortcuts (excerpted below), then screened by a second, video-blind LLM that tries to answer from the question and options alone; candidates it can solve with high confidence are discarded before RL rollout mining.

\begin{figure*}[t]
\begin{promptbox}{promptorange}{MCQ\_DESIGN\_SYSTEM\_PROMPT (excerpt)}
\begin{Verbatim}[fontsize=\scriptsize,breaklines=true,breakanywhere=true]
You are an expert multiple-choice question (MCQ) designer, skilled at creating high-quality questions based on video content.
Your task is to convert the given question-answer (QA) pairs into multiple-choice format (4 options A B C D).

## Design Principles

### 1. Option Length & Structure Balance (CRITICAL - HIGHEST PRIORITY)
Test-takers often guess by comparing option lengths.
- All 4 options MUST have nearly identical character/word counts (within 10-15% variance).
- The correct answer MUST NOT be the longest or the shortest option; if it is naturally long, EXTEND all distractors to match, if naturally short, SHORTEN it or extend distractors.
- Avoid "three short + one long" (or vice versa) patterns where the correct answer is the length outlier, or the only option with specific details/numbers.
- All options MUST use the same grammatical structure and level of specificity; if one option has a number/time/location, all should.

### 2. Distractor Design Principles
- Distractors must be related to elements that actually appear in the video, not fabricated, and should look plausible but be ruled out after careful viewing.
- Avoid absurd, common-sense-violating options; keep distractor type consistent with the correct answer (number/time/name/etc.).

### 3. Anti-Guessing Strategies
- Avoid extreme words ("always", "never", "all"); randomize the correct-answer position; do not make the correct answer more specific or vague than distractors.
- Question and option wording must not hint at the answer. Options MUST NOT be solvable by general world knowledge/common sense alone -- the question must require video-specific visual/audio evidence.
- The correct answer should not have any unique characteristic that sets it apart from distractors.

[Output JSON schema and a pre-submission length/structure self-check checklist omitted for brevity.]
\end{Verbatim}
\end{promptbox}

\begin{promptbox}{promptorange}{MCQ\_TEXT\_ONLY\_GUESSABILITY\_JUDGE}
\begin{Verbatim}[fontsize=\scriptsize,breaklines=true,breakanywhere=true]
You are a test analysis expert. Your task is to analyze whether a multiple-choice question about video content can be answered correctly using ONLY the text information (question and options), WITHOUT watching the actual video.

## Your Task
Given a multiple-choice question that was originally designed to test video understanding:
1. Analyze if the correct answer can be inferred from text alone.
2. If yes, identify what text-based clues allow this inference.
3. Provide your confidence level (0-100).

## Important Guidelines
- You CANNOT watch the video - you only have access to the question text and options.
- Be honest about whether you can infer the answer or not; if guessing randomly, say so with low confidence.

## Output Format
{"can_infer": true/false, "answer": "A/B/C/D/E", "confidence": 0-100, "reasoning_type": "option_length|option_structure|common_sense|option_content|question_hint|elimination|cannot_determine", "reasoning": "..."}

confidence: 0-30 random guess, 31-50 slight inclination, 51-70 reasonable inference, 71-100 high confidence from clear text clues.
reasoning_type: option_length/option_structure = options differ in length/grammar; common_sense = answerable from general knowledge; option_content = some options are obviously wrong; question_hint = question wording hints the answer; elimination = can narrow down by ruling out options; cannot_determine = no text-based way in.

## Critical Reminders
1. This is a VIDEO understanding question -- the "correct" answer is based on video content you cannot see.
2. Your goal is to detect if the question design has FLAWS that allow text-only inference.
3. If you can answer correctly with high confidence without the video, the question is poorly designed.
4. Be rigorous - don't claim you can infer if you are actually guessing.
\end{Verbatim}
\end{promptbox}
\end{figure*}

% \subsection{Tool-return prompt}

% \begin{promptbox}{promptorange}{ZOOM\_IN\_CLIP\_USER\_MESSAGE}
% \begin{Verbatim}[fontsize=\scriptsize,breaklines=true,breakanywhere=true]
% Zoom-in video clip [start, end] seconds:
% Use the provided zoom-in clip (and prior context) to output final <answer>...</answer>.
% Do not output <interval> in this turn.

% <zoom-in audio-video clip>
% \end{Verbatim}
% \end{promptbox}

%%%%%%%%%%%%%%%%%%%%%%%%%%%%%%%%%%%%%%%%%%%%%%%%%%%%%%%%%%%%

\end{document}

%% file: related_work.tex
\subsection{Long Omni Video Reasoning}

Long-form video understanding combines long-context perception, temporal localization, and multi-step reasoning. Recent work improves long-video reasoning by exposing temporal structure through explicit markers, grounding objectives, or time-aware video representations \citep{chen2024timemarkerversatilevideollmlong,Wu2024,zhang2026timelensrethinkingvideotemporal}. For audio-visual inputs, the challenge is sharper: decisive evidence may appear in vision, audio, or their temporal coincidence. Recent omni reasoning models therefore explore cross-modal cooperation and time-aware omni representations \citep{du2025crabunifiedaudiovisualscene,xing2025echoinkr1exploringaudiovisualreasoning,zhong2025omnir1reinforcementlearningomnimodal,chen2025chronusomniimprovingtimeawareness,xiao2025pixclip}. These works show that omni models need temporally structured perception, but most still process evidence under a fixed observation budget or optimize reasoning after evidence has already been selected. OmniReasoner instead focuses on the preceding decision: whether additional evidence is needed and where on the audio-video timeline it should be retrieved.

\subsection{Agentic Tool Use}

Another line treats long-video understanding as interactive evidence acquisition. Agentic and tool-based systems search, select, or revisit informative frames and segments \citep{wang2024videoagentlongformvideounderstanding,gao2026videotiraccurateunderstandinglong,fu2025lover1advancinglongvideo,pan2025timesearchradaptivetemporal,yang2025longvtincentivizingthinkinglong}, while video reasoning methods encourage thinking, reflection, verification, or frame spotlighting \citep{wang2025videothinkersparkingthinkingvideos,li2025revisortextualreflectionmultimodal,he2025framethinkerlearningthinklong,tang2025videor4reinforcingtextrichvideo, gao2026claimlevelrubricrewardsvideo}. These works support our motivation that long videos should be inspected adaptively rather than consumed uniformly. However, they largely define actions over visual evidence. OmniReasoner instead targets omni-modal tool-use post-training: it teaches the model whether and where to call a zoom-in tool for higher-fidelity audio-video evidence, with TimeAnchor keeping the tool's temporal argument valid across the sampling granularities observed before and after the call.

%% file: main.bib
@article{xu2025qwen3,
  title={Qwen3-omni technical report},
  author={Xu, Jin and Guo, Zhifang and Hu, Hangrui and Chu, Yunfei and Wang, Xiong and He, Jinzheng and Wang, Yuxuan and Shi, Xian and He, Ting and Zhu, Xinfa and others},
  journal={arXiv preprint arXiv:2509.17765},
  year={2025}
}

@article{ye2025omnivinci,
  title={OmniVinci: Enhancing Architecture and Data for Omni-Modal Understanding LLM},
  author={Ye, Hanrong and Yang, Chao-Han Huck and Goel, Arushi and Huang, Wei and Zhu, Ligeng and Su, Yuanhang and Lin, Sean and Cheng, An-Chieh and Wan, Zhen and Tian, Jinchuan and others},
  journal={arXiv preprint arXiv:2510.15870},
  year={2025}
}

@article{fu2025vita,
  title={Vita-1.5: Towards gpt-4o level real-time vision and speech interaction},
  author={Fu, Chaoyou and Lin, Haojia and Wang, Xiong and Zhang, Yi-Fan and Shen, Yunhang and Liu, Xiaoyu and Cao, Haoyu and Long, Zuwei and Gao, Heting and Li, Ke and others},
  journal={arXiv preprint arXiv:2501.01957},
  year={2025}
}

@article{dai2026latentomni,
  title={LatentOmni: Rethinking Omni-Modal Understanding via Unified Audio-Visual Latent Reasoning},
  author={Dai, Yifan and Wu, Zhenhua and Zeng, Bohan and Hua, Daili and Liu, Jialing and Li, Bozhou and Wang, Yuran and Tong, Chengzhuo and Liang, Hao and Ma, Xiaochen and others},
  journal={arXiv preprint arXiv:2605.22012},
  year={2026}
}

@misc{gao2026claimlevelrubricrewardsvideo,
      title={Claim-Level Rubric Rewards for Video Caption Reinforcement Learning}, 
      author={Mingqi Gao and Hongyuan Dong and Yifei Chen and Zhisheng Zhong and Zheng Ruan and Wenjin Hou and Yu Chen and Han Hu and Yansong Tang},
      year={2026},
      eprint={2607.05150},
      archivePrefix={arXiv},
      primaryClass={cs.CV},
      url={https://arxiv.org/abs/2607.05150}, 
}

@inproceedings{gao2017tall,
  title={Tall: Temporal activity localization via language query},
  author={Gao, Jiyang and Sun, Chen and Yang, Zhenheng and Nevatia, Ram},
  booktitle={Proceedings of the IEEE international conference on computer vision},
  pages={5267--5275},
  year={2017},
  doi={10.1109/ICCV.2017.563}
}

@article{fu2024video,
title={Video-MME: The First-Ever Comprehensive Evaluation Benchmark of Multi-modal LLMs in Video Analysis},
author={Fu, Chaoyou and Dai, Yuhan and Luo, Yongdong and Li, Lei and Ren, Shuhuai and Zhang, Renrui and Wang, Zihan and Zhou, Chenyu and Shen, Yunhang and Zhang, Mengdan and others},
journal={arXiv preprint arXiv:2405.21075},
year={2024}
}

@misc{Farré2024FineVideo,
  title={FineVideo},
  author={Farré, Miquel and Marafioti, Andi and Tunstall, Lewis and Von Werra, Leandro and Wolf, Thomas},
  year={2024},
  howpublished={\url{https://huggingface.co/datasets/HuggingFaceFV/finevideo}},
}

@misc{queen2025cgbenchbenchmarkinglanguagemodel,
      title={CGBench: Benchmarking Language Model Scientific Reasoning for Clinical Genetics Research}, 
      author={Owen Queen and Harrison G. Zhang and James Zou},
      year={2025},
      eprint={2510.11985},
      archivePrefix={arXiv},
      primaryClass={cs.AI},
      url={https://arxiv.org/abs/2510.11985}, 
}

@article{team2023gemini,
  title={Gemini: a family of highly capable multimodal models},
  author={Team, Gemini and Anil, Rohan and Borgeaud, Sebastian and Alayrac, Jean-Baptiste and Yu, Jiahui and Soricut, Radu and Schalkwyk, Johan and Dai, Andrew M and Hauth, Anja and Millican, Katie and others},
  journal={arXiv preprint arXiv:2312.11805},
  year={2023}
}

@article{tao2026lvomnibench,
  title={LVOmniBench: Pioneering Long Audio-Video Understanding Evaluation for Omnimodal LLMs},
  author={Tao, Keda and Zheng, Yuhua and Xu, Jia and Du, Wenjie and Shao, Kele and Wang, Hesong and Chen, Xueyi and Jin, Xin and Zhu, Junhan and Yu, Bohan and others},
  journal={arXiv preprint arXiv:2603.19217},
  year={2026}
}

@article{xiao2025pixclip,
  title={PixCLIP: Achieving Fine-grained Visual Language Understanding via Any-granularity Pixel-Text Alignment Learning},
  author={Xiao, Yicheng and Chen, Yu and Ma, Haoxuan and Hong, Jiale and Li, Caorui and Wu, Lingxiang and Guo, Haiyun and Wang, Jinqiao},
  journal={arXiv preprint arXiv:2511.04601},
  year={2025}
}

@misc{liu2025faithfulnessvisualthinkingmeasurement,
      title={On the Faithfulness of Visual Thinking: Measurement and Enhancement}, 
      author={Zujing Liu and Junwen Pan and Qi She and Yuan Gao and Guisong Xia},
      year={2025},
      eprint={2510.23482},
      archivePrefix={arXiv},
      primaryClass={cs.CV},
      url={https://arxiv.org/abs/2510.23482}, 
}

@inproceedings{abnar2020quantifying,
  title={Quantifying attention flow in transformers},
  author={Abnar, Samira and Zuidema, Willem},
  booktitle={Proceedings of the 58th annual meeting of the association for computational linguistics},
  pages={4190--4197},
  year={2020},
  doi={10.18653/v1/2020.acl-main.385}
}

@article{xu2026flowprune,
  title={FlowPrune: Accelerating Attention Flow Calculation by Pruning Flow Network},
  author={Xu, Shuo and Chen, Yu and Lin, Shuxia and Geng, Xin and Yang, Xu},
  journal={Advances in Neural Information Processing Systems},
  volume={38},
  pages={134236--134261},
  year={2025},
  url={https://proceedings.neurips.cc/paper_files/paper/2025/hash/c2cd7fa98591516c025d7632b2586ca1-Abstract-Conference.html}
}

@inproceedings{Zhang_2025,
   title={LMMs-Eval: Reality Check on the Evaluation of Large Multimodal Models},
   url={http://dx.doi.org/10.18653/v1/2025.findings-naacl.51},
   DOI={10.18653/v1/2025.findings-naacl.51},
   booktitle={Findings of the Association for Computational Linguistics: NAACL 2025},
   publisher={Association for Computational Linguistics},
   author={Zhang, Kaichen and Li, Bo and Zhang, Peiyuan and Pu, Fanyi and Cahyono, Joshua Adrian and Hu, Kairui and Liu, Shuai and Zhang, Yuanhan and Yang, Jingkang and Li, Chunyuan and Liu, Ziwei},
   year={2025},
   pages={881–916} }

@misc{sheng2024hybridflow,
      title={HybridFlow: A Flexible and Efficient RLHF Framework}, 
      author={Guangming Sheng and Chi Zhang and Zilingfeng Ye and Xibin Wu and Wang Zhang and Ru Zhang and Yanghua Peng and Haibin Lin and Chuan Wu},
      year={2024},
      eprint={2409.19256},
      archivePrefix={arXiv},
      primaryClass={cs.LG},
      url={https://arxiv.org/abs/2409.19256}, 
}

@misc{vonwerra2020trl,
      title={TRL: Transformers Reinforcement Learning}, 
      author={Leandro von Werra and Younes Belkada and Lewis Tunstall and Edward Beeching and Tristan Thrush and Nathan Lambert and Shengyi Huang and Kashif Rasul and Quentin Gallouedec},
      year={2020},
      publisher={GitHub},
      journal={GitHub repository},
      url={https://github.com/huggingface/trl}, 
}

@article{liu2025ola,
  title={Ola: Pushing the frontiers of omni-modal language model},
  author={Liu, Zuyan and Dong, Yuhao and Wang, Jiahui and Liu, Ziwei and Hu, Winston and Lu, Jiwen and Rao, Yongming},
  journal={arXiv preprint arXiv:2502.04328},
  year={2025}
}

@article{cheng2024videollama,
  title={Videollama 2: Advancing spatial-temporal modeling and audio understanding in video-llms},
  author={Cheng, Zesen and Leng, Sicong and Zhang, Hang and Xin, Yifei and Li, Xin and Chen, Guanzheng and Zhu, Yongxin and Zhang, Wenqi and Luo, Ziyang and Zhao, Deli and others},
  journal={arXiv preprint arXiv:2406.07476},
  year={2024}
}

@misc{cheng2025videoholmes,
      title={Video-Holmes: Can MLLM Think Like Holmes for Complex Video Reasoning?},
      author={Junhao Cheng and Yuying Ge and Teng Wang and Yixiao Ge and Jing Liao and Ying Shan},
      year={2025},
      eprint={2505.21374},
      archivePrefix={arXiv},
      primaryClass={cs.CV},
      url={https://arxiv.org/abs/2505.21374},
}

@article{cui2026minicpm,
  title={MiniCPM-o 4.5: Towards Real-Time Full-Duplex Omni-Modal Interaction},
  author={Cui, Junbo and Xu, Bokai and Wang, Chongyi and Yu, Tianyu and Sun, Weiyue and Xu, Yingjing and Wang, Tianran and He, Zhihui and Ma, Wenshuo and Cai, Tianchi and others},
  journal={arXiv preprint arXiv:2604.27393},
  year={2026}
}

@misc{hong2026worldsenseevaluatingrealworldomnimodal,
      title={WorldSense: Evaluating Real-world Omnimodal Understanding for Multimodal LLMs}, 
      author={Jack Hong and Shilin Yan and Jiayin Cai and Xiaolong Jiang and Yao Hu and Weidi Xie},
      year={2026},
      eprint={2502.04326},
      archivePrefix={arXiv},
      primaryClass={cs.CV},
      url={https://arxiv.org/abs/2502.04326}, 
}

@misc{shao2024deepseekmathpushinglimitsmathematical,
      title={DeepSeekMath: Pushing the Limits of Mathematical Reasoning in Open Language Models}, 
      author={Zhihong Shao and Peiyi Wang and Qihao Zhu and Runxin Xu and Junxiao Song and Xiao Bi and Haowei Zhang and Mingchuan Zhang and Y. K. Li and Y. Wu and Daya Guo},
      year={2024},
      eprint={2402.03300},
      archivePrefix={arXiv},
      primaryClass={cs.CL},
      url={https://arxiv.org/abs/2402.03300}, 
}

@misc{deepseekai2025deepseekv32pushingfrontieropen,
      title={DeepSeek-V3.2: Pushing the Frontier of Open Large Language Models}, 
      author={DeepSeek-AI and Aixin Liu and Aoxue Mei and Bangcai Lin and Bing Xue and Bingxuan Wang and Bingzheng Xu and Bochao Wu and Bowei Zhang and Chaofan Lin and Chen Dong and Chengda Lu and Chenggang Zhao and Chengqi Deng and Chenhao Xu and Chong Ruan and Damai Dai and Daya Guo and Dejian Yang and Deli Chen and Erhang Li and Fangqi Zhou and Fangyun Lin and Fucong Dai and Guangbo Hao and Guanting Chen and Guowei Li and H. Zhang and Hanwei Xu and Hao Li and Haofen Liang and Haoran Wei and Haowei Zhang and Haowen Luo and Haozhe Ji and Honghui Ding and Hongxuan Tang and Huanqi Cao and Huazuo Gao and Hui Qu and Hui Zeng and Jialiang Huang and Jiashi Li and Jiaxin Xu and Jiewen Hu and Jingchang Chen and Jingting Xiang and Jingyang Yuan and Jingyuan Cheng and Jinhua Zhu and Jun Ran and Junguang Jiang and Junjie Qiu and Junlong Li and Junxiao Song and Kai Dong and Kaige Gao and Kang Guan and Kexin Huang and Kexing Zhou and Kezhao Huang and Kuai Yu and Lean Wang and Lecong Zhang and Lei Wang and Liang Zhao and Liangsheng Yin and Lihua Guo and Lingxiao Luo and Linwang Ma and Litong Wang and Liyue Zhang and M. S. Di and M. Y Xu and Mingchuan Zhang and Minghua Zhang and Minghui Tang and Mingxu Zhou and Panpan Huang and Peixin Cong and Peiyi Wang and Qiancheng Wang and Qihao Zhu and Qingyang Li and Qinyu Chen and Qiushi Du and Ruiling Xu and Ruiqi Ge and Ruisong Zhang and Ruizhe Pan and Runji Wang and Runqiu Yin and Runxin Xu and Ruomeng Shen and Ruoyu Zhang and S. H. Liu and Shanghao Lu and Shangyan Zhou and Shanhuang Chen and Shaofei Cai and Shaoyuan Chen and Shengding Hu and Shengyu Liu and Shiqiang Hu and Shirong Ma and Shiyu Wang and Shuiping Yu and Shunfeng Zhou and Shuting Pan and Songyang Zhou and Tao Ni and Tao Yun and Tian Pei and Tian Ye and Tianyuan Yue and Wangding Zeng and Wen Liu and Wenfeng Liang and Wenjie Pang and Wenjing Luo and Wenjun Gao and Wentao Zhang and Xi Gao and Xiangwen Wang and Xiao Bi and Xiaodong Liu and Xiaohan Wang and Xiaokang Chen and Xiaokang Zhang and Xiaotao Nie and Xin Cheng and Xin Liu and Xin Xie and Xingchao Liu and Xingkai Yu and Xingyou Li and Xinyu Yang and Xinyuan Li and Xu Chen and Xuecheng Su and Xuehai Pan and Xuheng Lin and Xuwei Fu and Y. Q. Wang and Yang Zhang and Yanhong Xu and Yanru Ma and Yao Li and Yao Li and Yao Zhao and Yaofeng Sun and Yaohui Wang and Yi Qian and Yi Yu and Yichao Zhang and Yifan Ding and Yifan Shi and Yiliang Xiong and Ying He and Ying Zhou and Yinmin Zhong and Yishi Piao and Yisong Wang and Yixiao Chen and Yixuan Tan and Yixuan Wei and Yiyang Ma and Yiyuan Liu and Yonglun Yang and Yongqiang Guo and Yongtong Wu and Yu Wu and Yuan Cheng and Yuan Ou and Yuanfan Xu and Yuduan Wang and Yue Gong and Yuhan Wu and Yuheng Zou and Yukun Li and Yunfan Xiong and Yuxiang Luo and Yuxiang You and Yuxuan Liu and Yuyang Zhou and Z. F. Wu and Z. Z. Ren and Zehua Zhao and Zehui Ren and Zhangli Sha and Zhe Fu and Zhean Xu and Zhenda Xie and Zhengyan Zhang and Zhewen Hao and Zhibin Gou and Zhicheng Ma and Zhigang Yan and Zhihong Shao and Zhixian Huang and Zhiyu Wu and Zhuoshu Li and Zhuping Zhang and Zian Xu and Zihao Wang and Zihui Gu and Zijia Zhu and Zilin Li and Zipeng Zhang and Ziwei Xie and Ziyi Gao and Zizheng Pan and Zongqing Yao and Bei Feng and Hui Li and J. L. Cai and Jiaqi Ni and Lei Xu and Meng Li and Ning Tian and R. J. Chen and R. L. Jin and S. S. Li and Shuang Zhou and Tianyu Sun and X. Q. Li and Xiangyue Jin and Xiaojin Shen and Xiaosha Chen and Xinnan Song and Xinyi Zhou and Y. X. Zhu and Yanping Huang and Yaohui Li and Yi Zheng and Yuchen Zhu and Yunxian Ma and Zhen Huang and Zhipeng Xu and Zhongyu Zhang and Dongjie Ji and Jian Liang and Jianzhong Guo and Jin Chen and Leyi Xia and Miaojun Wang and Mingming Li and Peng Zhang and Ruyi Chen and Shangmian Sun and Shaoqing Wu and Shengfeng Ye and T. Wang and W. L. Xiao and Wei An and Xianzu Wang and Xiaowen Sun and Xiaoxiang Wang and Ying Tang and Yukun Zha and Zekai Zhang and Zhe Ju and Zhen Zhang and Zihua Qu},
      year={2025},
      eprint={2512.02556},
      archivePrefix={arXiv},
      primaryClass={cs.CL},
      url={https://arxiv.org/abs/2512.02556}, 
}

@misc{wang2025videothinkersparkingthinkingvideos,
      title={Video-Thinker: Sparking "Thinking with Videos" via Reinforcement Learning}, 
      author={Shijian Wang and Jiarui Jin and Xingjian Wang and Linxin Song and Runhao Fu and Hecheng Wang and Zongyuan Ge and Yuan Lu and Xuelian Cheng},
      year={2025},
      eprint={2510.23473},
      archivePrefix={arXiv},
      primaryClass={cs.CV},
      url={https://arxiv.org/abs/2510.23473}, 
}

@misc{yang2025longvtincentivizingthinkinglong,
      title={LongVT: Incentivizing "Thinking with Long Videos" via Native Tool Calling}, 
      author={Zuhao Yang and Sudong Wang and Kaichen Zhang and Keming Wu and Sicong Leng and Yifan Zhang and Bo Li and Chengwei Qin and Shijian Lu and Xingxuan Li and Lidong Bing},
      year={2025},
      eprint={2511.20785},
      archivePrefix={arXiv},
      primaryClass={cs.CV},
      url={https://arxiv.org/abs/2511.20785}, 
}

@misc{fu2025lover1advancinglongvideo,
      title={LOVE-R1: Advancing Long Video Understanding with an Adaptive Zoom-in Mechanism via Multi-Step Reasoning},
      author={Shenghao Fu and Qize Yang and Yuan-Ming Li and others},
      year={2025},
      eprint={2509.24786},
      archivePrefix={arXiv},
      primaryClass={cs.CV},
      url={https://arxiv.org/abs/2509.24786},
}

@misc{li2025revisortextualreflectionmultimodal,
      title={REVISOR: Beyond Textual Reflection, Towards Multimodal Introspective Reasoning in Long-Form Video Understanding}, 
      author={Jiaze Li and Hao Yin and Wenhui Tan and Jingyang Chen and Boshen Xu and Yuxun Qu and Yijing Chen and Jianzhong Ju and Zhenbo Luo and Jian Luan},
      year={2025},
      eprint={2511.13026},
      archivePrefix={arXiv},
      primaryClass={cs.CV},
      url={https://arxiv.org/abs/2511.13026}, 
}

@misc{chen2024timemarkerversatilevideollmlong,
      title={TimeMarker: A Versatile Video-LLM for Long and Short Video Understanding with Superior Temporal Localization Ability}, 
      author={Shimin Chen and Xiaohan Lan and Yitian Yuan and Zequn Jie and Lin Ma},
      year={2024},
      eprint={2411.18211},
      archivePrefix={arXiv},
      primaryClass={cs.CV},
      url={https://arxiv.org/abs/2411.18211}, 
}

@misc{zhang2026timelensrethinkingvideotemporal,
      title={TimeLens: Rethinking Video Temporal Grounding with Multimodal LLMs}, 
      author={Jun Zhang and Teng Wang and Yuying Ge and Yixiao Ge and Xinhao Li and Ying Shan and Limin Wang},
      year={2026},
      eprint={2512.14698},
      archivePrefix={arXiv},
      primaryClass={cs.CV},
      url={https://arxiv.org/abs/2512.14698}, 
}

@misc{pan2025timesearchradaptivetemporal,
      title={TimeSearch-R: Adaptive Temporal Search for Long-Form Video Understanding via Self-Verification Reinforcement Learning},
      author={Junwen Pan and Qizhe Zhang and Rui Zhang and others},
      year={2025},
      eprint={2511.05489},
      archivePrefix={arXiv},
      primaryClass={cs.CV},
      url={https://arxiv.org/abs/2511.05489},
}

@misc{tang2025videor4reinforcingtextrichvideo,
      title={Video-R4: Reinforcing Text-Rich Video Reasoning with Visual Rumination}, 
      author={Yolo Y. Tang and Daiki Shimada and Hang Hua and Chao Huang and Jing Bi and Rogerio Feris and Chenliang Xu},
      year={2025},
      eprint={2511.17490},
      archivePrefix={arXiv},
      primaryClass={cs.CV},
      url={https://arxiv.org/abs/2511.17490}, 
}

@misc{he2025framethinkerlearningthinklong,
      title={FrameThinker: Learning to Think with Long Videos via Multi-Turn Frame Spotlighting}, 
      author={Zefeng He and Xiaoye Qu and Yafu Li and Siyuan Huang and Daizong Liu and Yu Cheng},
      year={2025},
      eprint={2509.24304},
      archivePrefix={arXiv},
      primaryClass={cs.CV},
      url={https://arxiv.org/abs/2509.24304}, 
}

@article{Wu2024,
  title = {Number it: Temporal Grounding Videos like Flipping Manga},
  author = {Yongliang Wu and Xinting Hu and Yuyang Sun et al.},
  journal = {arXiv preprint arXiv:2411.10332},
  year = {2024},
  url = {https://arxiv.org/abs/2411.10332}
}

@misc{wang2024videoagentlongformvideounderstanding,
      title={VideoAgent: Long-form Video Understanding with Large Language Model as Agent}, 
      author={Xiaohan Wang and Yuhui Zhang and Orr Zohar and Serena Yeung-Levy},
      year={2024},
      eprint={2403.10517},
      archivePrefix={arXiv},
      primaryClass={cs.CV},
      url={https://arxiv.org/abs/2403.10517}, 
}

@misc{gao2026videotiraccurateunderstandinglong,
      title={VideoTIR: Accurate Understanding for Long Videos with Efficient Tool-Integrated Reasoning}, 
      author={Zhe Gao and Shiyu Shen and Taifeng Chai and Weinong Wang and Haotian Xu and Xing W and Wenbin Li and Qi Fan and Yang Gao and Dacheng Tao},
      year={2026},
      eprint={2603.25021},
      archivePrefix={arXiv},
      primaryClass={cs.CV},
      url={https://arxiv.org/abs/2603.25021}, 
}

@misc{li2026omnivideobench,
      title={OmniVideoBench: Towards Audio-Visual Understanding Evaluation for Omni MLLMs},
      author={Caorui Li and Yu Chen and Yiyan Ji and Jin Xu and Zhenyu Cui and Shihao Li and Yuanxing Zhang and Wentao Wang and Zhenghao Song and Dingling Zhang and Ying He and Haoxiang Liu and Yuxuan Wang and Qiufeng Wang and Jiafu Tang and Zhenhe Wu and Jiehui Luo and Zhiyu Pan and Weihao Xie and Chenchen Zhang and Zhaohui Wang and Jiayi Tian and Yanghai Wang and Zhe Cao and Minxin Dai and Ke Wang and Runzhe Wen and Yinghao Ma and Yaning Pan and Sungkyun Chang and Termeh Taheri and Haiwen Xia and Christos Plachouras and Emmanouil Benetos and Yizhi Li and Ge Zhang and Jian Yang and Tianhao Peng and Zili Wang and Minghao Liu and Junran Peng and Zhaoxiang Zhang and Jiaheng Liu},
      year={2026},
      eprint={2510.10689},
      archivePrefix={arXiv},
      primaryClass={cs.AI},
      url={https://arxiv.org/abs/2510.10689},
}

@misc{zhou2026dailyomni,
      title={Daily-Omni: Towards Audio-Visual Reasoning with Temporal Alignment across Modalities},
      author={Ziwei Zhou and Rui Wang and Zuxuan Wu and Yu-Gang Jiang},
      year={2026},
      eprint={2505.17862},
      archivePrefix={arXiv},
      primaryClass={cs.AI},
      url={https://arxiv.org/abs/2505.17862},
}

@misc{chen2025chronusomniimprovingtimeawareness,
      title={ChronusOmni: Improving Time Awareness of Omni Large Language Models}, 
      author={Yijing Chen and Yihan Wu and Kaisi Guan and Yuchen Ren and Yuyue Wang and Ruihua Song and Liyun Ru},
      year={2025},
      eprint={2512.09841},
      archivePrefix={arXiv},
      primaryClass={cs.CL},
      url={https://arxiv.org/abs/2512.09841}, 
}

@misc{chen2026omnivideor1reinforcingaudiovisual,
      title={OmniVideo-R1: Reinforcing Audio-visual Reasoning with Query Intention and Modality Attention},
      author={Zhangquan Chen and Jiale Tao and Ruihuang Li and others},
      year={2026},
      eprint={2602.05847},
      archivePrefix={arXiv},
      primaryClass={cs.CV},
      url={https://arxiv.org/abs/2602.05847},
}

@misc{li2026omnigaiatowardsnativeomnimodal,
      title={OmniGAIA: Towards Native Omni-Modal AI Agents},
      author={Xiaoxi Li and Wenxiang Jiao and Jiarui Jin and others},
      year={2026},
      eprint={2602.22897},
      archivePrefix={arXiv},
      primaryClass={cs.AI},
      url={https://arxiv.org/abs/2602.22897},
}

@misc{du2025crabunifiedaudiovisualscene,
      title={Crab: A Unified Audio-Visual Scene Understanding Model with Explicit Cooperation}, 
      author={Henghui Du and Guangyao Li and Chang Zhou and Chunjie Zhang and Alan Zhao and Di Hu},
      year={2025},
      eprint={2503.13068},
      archivePrefix={arXiv},
      primaryClass={cs.CV},
      url={https://arxiv.org/abs/2503.13068}, 
}

@misc{xing2025echoinkr1exploringaudiovisualreasoning,
      title={EchoInk-R1: Exploring Audio-Visual Reasoning in Multimodal LLMs via Reinforcement Learning}, 
      author={Zhenghao Xing and Xiaowei Hu and Chi-Wing Fu and Wenhai Wang and Jifeng Dai and Pheng-Ann Heng},
      year={2025},
      eprint={2505.04623},
      archivePrefix={arXiv},
      primaryClass={cs.CV},
      url={https://arxiv.org/abs/2505.04623}, 
}

@misc{zhong2025omnir1reinforcementlearningomnimodal,
      title={Omni-R1: Reinforcement Learning for Omnimodal Reasoning via Two-System Collaboration},
      author={Hao Zhong and Muzhi Zhu and Zongze Du and others},
      year={2025},
      eprint={2505.20256},
      archivePrefix={arXiv},
      primaryClass={cs.AI},
      url={https://arxiv.org/abs/2505.20256},
}
